\documentclass[sn-mathphys]{sn-jnl}
 
\jyear{2025}

\usepackage{amsfonts}
\usepackage{amsmath}
\usepackage{booktabs}
\usepackage{multirow}
\usepackage{amssymb}
\usepackage{bbding}
\usepackage{pifont}
\usepackage{wasysym}
\usepackage{color}
\usepackage{siunitx}
\usepackage{graphicx}
\usepackage{diagbox}
\usepackage{textcomp}
\usepackage{siunitx}
\usepackage{caption}
\usepackage{subcaption}
\usepackage{mdframed}  
\usepackage{xcolor}
\usepackage{listings}
\usepackage{longtable}
\usepackage{ragged2e}
\usepackage{xspace}
\usepackage{tabularx}

\theoremstyle{thmstyletwo}

\theoremstyle{thmstylethree}

\usepackage{graphicx}  

\usepackage{xurl}
\usepackage{hyperref}
\usepackage{url}
\usepackage{booktabs}
\usepackage{algpseudocode}
\usepackage{enumerate}
\usepackage[english]{babel}
\usepackage{blindtext}
\usepackage{subcaption}
\usepackage{bm}
\usepackage{xspace}
\usepackage{multirow}
\usepackage{threeparttable}
\usepackage{float}
\usepackage{flafter}
\usepackage{comment}

\usepackage{ulem}
\usepackage{verbatim}
\usepackage{array}
\usepackage{enumitem}
\usepackage{booktabs}
\usepackage{multirow}
\usepackage{colortbl}
\usepackage{lipsum}
\usepackage{wrapfig}
\usepackage{pifont}
\usepackage{subcaption}

\begin{document}


\title[AI4SoS]{AI-Driven Automation Can Become the Foundation of Next-Era Science of Science Research}

\author{
  Renqi Chen\textsuperscript{1,}\footnote{Equal contributions.},
  Haoyang Su\textsuperscript{1,}\footnotemark[1],
  Shixiang Tang\textsuperscript{1,6}, 
  Zhenfei Yin\textsuperscript{2},
  Qi Wu\textsuperscript{3}, \quad
  Hui Li\textsuperscript{3}, \quad
  Ye Sun\textsuperscript{4},\quad
  Nanqing Dong\textsuperscript{1,5,}\footnote{Corresponding authors: Nanqing Dong (dongnanqing@pjlab.org.cn).},\\
  Wanli Ouyang\textsuperscript{1,6},\quad
  Philip Torr\textsuperscript{2}
}

\affil[1]{Shanghai Artificial Intelligence Laboratory}
\affil[2]{Department of Engineering Science, University of Oxford}
\affil[3]{Shanghai Institute for Science of Science}
\affil[4]{School of Mathematics, Southeast University}
\affil[5]{Shanghai Innovation Institute}
\affil[6]{Department of Information Engineering, Chinese University of Hong Kong}

\abstract{
The Science of Science (SoS) explores the mechanisms underlying scientific discovery, and offers valuable insights for enhancing scientific efficiency and fostering innovation. Traditional approaches often rely on simplistic assumptions and basic statistical tools, such as linear regression and rule-based simulations, which struggle to capture the complexity and scale of modern research ecosystems. {The advent of artificial intelligence (AI) presents a transformative opportunity for the next generation of SoS, enabling the automation of large-scale pattern discovery and uncovering insights previously unattainable. This paper offers a forward-looking perspective on the integration of Science of Science with AI for automated research pattern discovery and highlights key open challenges that could greatly benefit from AI.} We outline the advantages of AI over traditional methods, discuss potential limitations, and propose pathways to overcome them. Additionally, we present a preliminary multi-agent system as an illustrative example to simulate research societies, showcasing AI's ability to replicate real-world research patterns and accelerate progress in Science of Science research.
}

\maketitle

\addtocontents{toc}{\protect\setcounter{tocdepth}{0}} 

\section{Introduction}
Science of Science (SoS), a pivotal and rapidly evolving field, serves as a strategic compass for guiding the trajectory of scientific and technological progress. By analyzing the complex dynamics of research collaboration and scientific output across geographic and temporal scales, it sheds light on the factors that drive creativity and the emergence of scientific discoveries, with the goal of developing tools and policies to accelerate scientific advancement~\cite{fortunato2018science}.
Unlike broader social sciences that examine societal structures, SoS delves deep into the mechanisms that fuel scientific breakthroughs~\cite{bettencourt2009scientific,shi2015weaving,klavans2017type}—illuminating the hidden forces that propel discovery and transformation. Ultimately, SoS underscores that groundbreaking advancements are not solely the result of talented minds and quality data, but are profoundly shaped by effective resource allocation, supportive policies and well-designed organizational structures~\cite{wang2020science,wapman2022quantifying}.

\begin{figure}[t]
    \centering
\includegraphics[width=\linewidth]{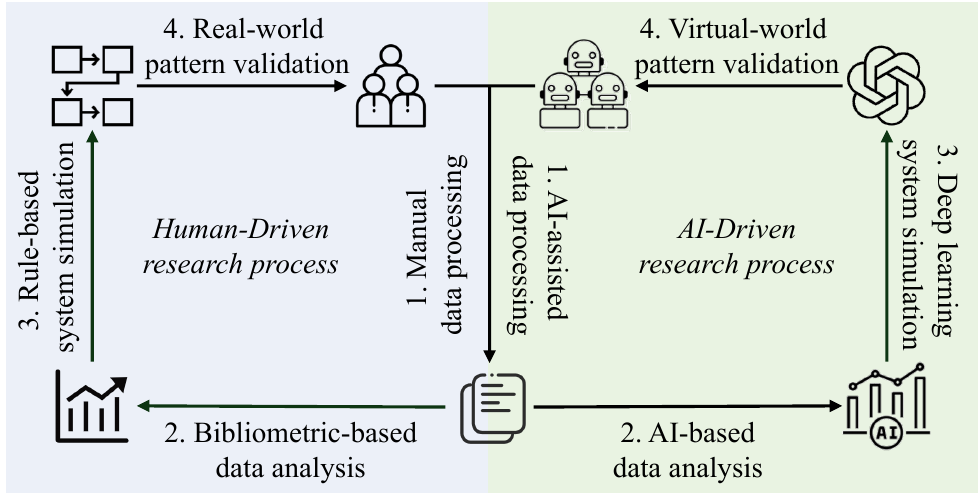}
    \caption{An illustration comparing human-driven and AI-driven research processes in the SoS, highlighting step-by-step differences across four key stages in order: \textit{data processing}, \textit{data analysis}, \textit{system simulation}, and \textit{pattern validation}.}
    \label{fig:intro}
\end{figure}

In recent years, the deep fusion of AI and SoS has become more feasible and promising than ever before. First, the increasing availability of large-scale scholarly data—publications, funding records, and collaboration networks—provides unprecedented opportunities to gain deeper insights into the evolution of scientific progress. Second, rapid advancements in AI technologies, such as large language models (LLMs), along with improvements in computational power, have greatly enhanced our ability to analyze and interpret complex scientific information with unprecedented accuracy and scale. These technological breakthroughs mark a critical moment for integrating AI into SoS, paving the way for a more data-driven approach to understanding and guiding research pattern discovery. {While some recent works have begun exploring autonomous scientific discovery, the field remains in its infancy, and there is still much progress to be made before realizing its full potential.}

{In this paper, we take a step forward by providing the first glimpse into the integration of AI and SoS for automated research pattern discovery. \textbf{We take the position that AI has the potential to revolutionize SoS, enabling the next generation of research by not only automating traditional research processes but also providing a sandbox for SoS research, allowing scientists to observe research processes in action and validate their hypotheses.}}
As illustrated in Fig.~\ref{fig:intro}, traditional SoS methods have primarily relied on manual data processing, bibliometric-based data analysis, rule-based system simulations, and real-world pattern validation. In contrast, AI-driven SoS leverages automated techniques to assist scientists in processing and analyzing data while offering more advanced and comprehensive systems for simulation and validation. This shift from human-driven to AI-driven methodologies unlocks the potential for more efficient, scalable, and data-driven analysis, ultimately providing deeper and more actionable insights into the mechanisms that shape scientific progress.
Thus, we define AI for SoS (AI4SoS) as a cross-disciplinary field that not only focuses on facilitating each step within the research process but also aims to achieve fully automated SoS research to uncover the hidden forces driving scientific innovation. This distinguishes AI4SoS from existing AI for Science approaches, which focus on using AI tools to solve domain-specific scientific problems~\cite{gao2024drugclip,abramson2024accurate,chang2024bidirectional}.

To consolidate our insights, we propose a forward-looking hierarchy of different levels of AI4SoS automation in Sec.~\ref{sec:levels}, which outlines a possible step-by-step approach to achieving the goal of fully automated SoS discovery. Within each level, we describe the key differences compared to previous levels and provide related examples. 
In Sec.~\ref{sec:advantage}, we highlight critical open problems in SoS where AI offers advantages.
Despite its promise, we discuss challenges such as data imbalance across disciplines in Sec.~\ref{sec:challenges}, overwhelming parameters in the simulation system of scientific societies, and the need for a reasonable evaluation system to validate the reliability of the simulation. We also propose possible pathways to overcome these challenges. Last but not least, we introduce a preliminary multi-agent system to simulate research societies in Sec.~\ref{sec:experiment}, illustrating AI’s capability to enable fully automated pattern discovery.

\begin{table*}[t]
\centering
\begin{threeparttable}
\caption{Comparison between AI for Science and AI for Science of Science.}
\label{tab:ai_comparison}
\begin{tabularx}{\textwidth}{X >{\raggedright\arraybackslash}p{0.37\textwidth} >{\raggedright\arraybackslash}p{0.38\textwidth}}
\toprule
Feature & AI for Science & AI for Science of Science \\
\midrule
\textbf{Focus} & Solving domain-specific scientific problems. & Understanding mechanisms of scientific progress. \\

\textbf{Approach} & Direct application of AI to address scientific challenges. & Meta-level analysis to enhance the research process. \\

\textbf{Examples} & Predicting weather, designing new drugs, optimizing materials. & Studying research collaboration trends, analyzing innovation triggers, mapping knowledge growth. \\
\bottomrule
\end{tabularx}
\end{threeparttable}
\end{table*}

\section{AI for Science of Science}
\subsection{Definition}

AI for SoS (AI4SoS) refers to the application of AI techniques to analyze, simulate, and validate the pattern of scientific research. 
It aims to leverage AI to study key aspects of the scientific ecosystem, including research productivity (e.g. individual published paper count), 
citation pattern (e.g. frequency and manner of citations),
collaboration network (e.g. interdisciplinary research collaboration), 
and the factors driving the advancement of scientific knowledge (e.g. funding and policy). 
Specifically, AI can drive the research process of SoS by automatically applying methods such as machine learning, data mining, and computational simulations, thereby uncovering scientific patterns.

\subsection{Comparison between AI for Science and AI4SoS}
Both AI for Science (AI4S) and AI4SoS aim to leverage AI to solve scientific problems. However, they differ in research goals. AI4S focuses on solving a particular scientific problem directly, such as weather prediction, drug discovery, and materials science. AI4SoS takes a meta-level approach, focusing on understanding the mechanisms of scientific progress to facilitate and accelerate research. More specifically, AI4SoS focuses on factors such as innovation drivers, research collaboration patterns, and the evolution of scientific knowledge, which are not bound to a single scientific problem.

\begin{figure*}[t]
    \centering
\includegraphics[width=\linewidth]{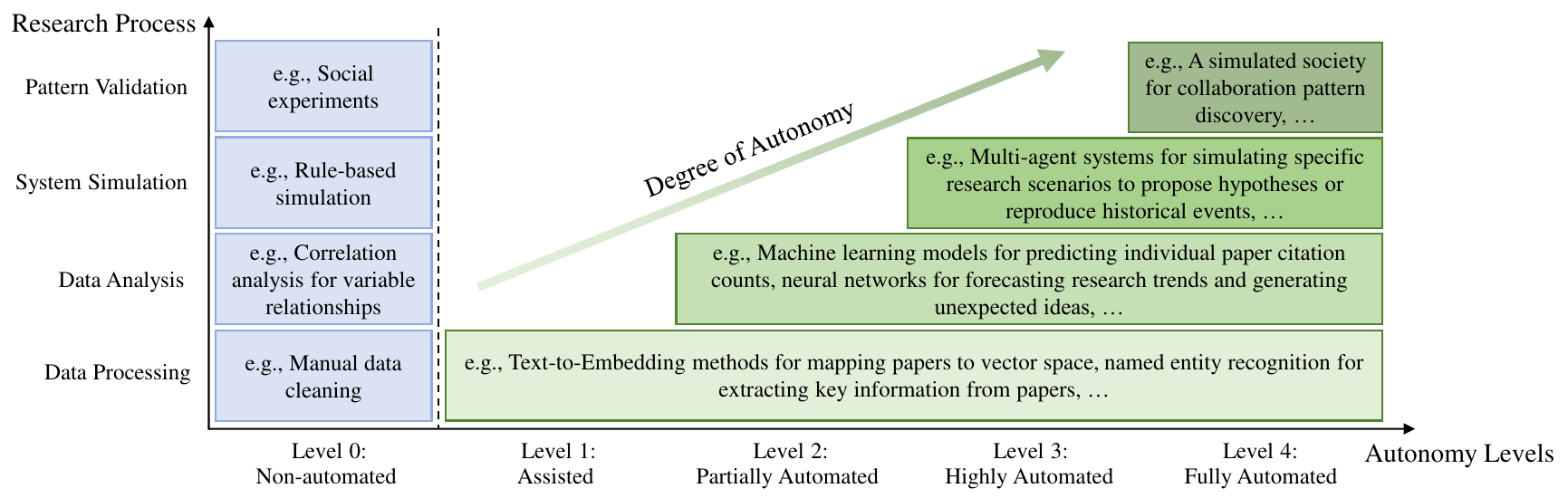}
    \caption{An overview of the five progressively advancing levels of autonomy in AI4SoS, with more green areas indicating that higher levels correspond to greater degrees of autonomy. Current research is primarily at Level 2 or below, with very limited work at Level 3, while fully automated SoS discovery remains in the prospective stage.}
    \label{fig:levels}
\end{figure*}

\subsection{Hierarchy of Automation Degree in AI4SoS}
The integration of AI techniques into scientific research follows a progressive hierarchy, reflecting the increasing autonomy and sophistication of AI systems in advancing the SoS field. As illustrated in Fig.~\ref{fig:levels}, we define five levels of autonomy, ranging from no AI involvement in pattern recognition and analysis to full autonomy in uncovering new scientific insights and guiding research strategies.

\noindent \textbf{Level 0: Non-automated SoS Discovery}
\label{sec:levels}
At this level, scientific pattern discovery is entirely human-driven and relies on traditional statistical methods. Researchers apply fundamental techniques such as probabilistic models, linear regression, and hypothesis testing to analyze scientific data and uncover patterns. AI is not involved in the process, and all tasks are conducted manually using well-established statistical procedures. Notable studies in this domain include the application of regression analysis to identify research trends~\citep{rzhetsky2015choosing}, correlation analysis to examine relationships between variables~\citep{alshebli2018preeminence}, and statistical estimation methods to explain observed scientific phenomena~\citep{liu2018hot, yin2019quantifying}.


\noindent \textbf{Level 1: AI-Assisted SoS Discovery}
In Level 1, AI only supports scientific data processing.
Specifically, AI methods are able to transform real-world scientific data into a more comprehensible form, including tasks such as completing and structuring bibliometric data, extracting key features such as author networks and institutional collaborations, and converting text information (e.g., papers, scientists) into embedding representations, thereby enhancing the efficiency and accuracy of data handling. 
However, AI's role remains supplementary, with human researchers still conducting data analysis, understanding and prediction.
From the perspective of AI4SoS, some related works include: utilizing text-to-embedding methods for mapping papers to vector space~\citep{shi2023surprising}, extracting key information from papers using named entity recognition~\citep{weston2019named}, and constructing networks for faculty mobility~\citep{clauset2015systematic}.

\noindent \textbf{Level 2: Partially Automated SoS Discovery}
In Level 2, AI techniques (e.g., supervised learning), play a central role in analyzing scientific data, enabling tasks such as predicting emerging trends, research hotspots and collaboration opportunities, based on historical patterns. This marks a shift from AI-assisted data processing to AI-driven data analysis. 
However, in this level, AI struggles to design and implement experiments automatically. For instance, a simulation environment that can automatically conduct scientific experiments is not available, therefore it is difficult to model hidden dynamic processes within the scientific ecosystem. 
Related works include the use of machine learning models to predict individual paper citation counts~\citep{xiao2016modeling}, neural networks for forecasting research trends and generating novel ideas~\citep{krenn2020predicting}, clustering publications based on citation relationships~\citep{van2017citation}, and applying structural topic models to extract topics from scientific texts~\citep{hofstra2020diversity}.


\noindent \textbf{Level 3: Highly Automated SoS Discovery}
In Level 3, AI not only drives the analysis but also designs and implements experiments to simulate scientific patterns in the real world.
In this case, researchers can compare results generated by simulation systems and those in the real world to explore strategies in SoS for potential real-world applications. 
While AI can support automatic experiment conduction, human supervision is required to define the specific application scenarios and corresponding experimental parameters (e.g., scientist information, boundary conditions) based on system feedback. Consequently, the authenticity and rationality of the system depends on whether the researchers have considered all relevant factors, making the automatic pattern validation difficult.
Research at this level is still in its early stages, including systems simulating specific research scenarios to propose hypotheses~\citep{ghafarollahi2024sciagents}, AI predicting outcomes under different simulation conditions to provide insights into collaboration patterns~\citep{su2024two}. and systems reproducing historical events based on specific environmental settings~\citep{yang2024oasis}.

\noindent \textbf{Level 4: Fully Automated SoS Discovery}
Level 4, the ultimate stage, represents complete automatic discovery in SoS. An AI-based virtual research society is conducted for end-to-end SoS discovery, including pattern analysis, prediction, and validation. 
Compared to systems in Level 3, systems in Level 4 function with continuous AI-based feedback loops to autonomously assess research plans and results to dynamically adjust parameters such as experimental settings, enabling virtual-world pattern validation as an alternative to real-world social experiments that may be aggressive.
At this stage, novel scientific insights can be uncovered without human intervention, and systems can adapt to new data and incorporate new insights in real time. Furthermore, ethical and governance frameworks are embedded, aligning the system’s actions with established guidelines for scientific integrity and accountability. 

Currently, most research remains at Level 2 or below, with only limited progress observed at Level 3, while fully automated SoS discovery is still in the exploratory stage. Looking ahead, several potential tasks are envisioned, including automated discovery of new collaboration patterns within the simulated scientific community~\citep{su2024two}, systems capable of simulating and conducting experiments in real-world settings~\citep{li2024ai}, and AI that continuously refines research directions based on emerging data~\citep{ofosu2024artificial}.

\section{Advantages of Automatic SoS Discovery} \label{sec:advantage}
In this section, we delve into critical open problems within the SoS that stand to benefit substantially from AI-driven automation. These problems are categorized into two primary areas: \textit{Forecasting Trends in Technology and Innovation} and \textit{Understanding the Dynamics of Research Society}. For each subproblem, we provide a brief background and outline key opportunities where AI offers advantages.
\subsection{Forecasting Trends in Technology and Innovation}
\subsubsection{Background of Problem}
Accurately forecasting the trajectory of science and technology is a crucial aspect of SoS, as it informs decisions related to funding, policy-making, and research prioritization. Two major challenges are predicting technological trends and identifying interdisciplinary opportunities.

\noindent \textbf{The Trend in Technological Development}~
Technological development follows intricate and often non-linear trajectories, making prediction difficult. To predict these trends, it is essential to understand which technologies are gaining momentum, identify emerging breakthroughs, and anticipate when they will transition from research to real-world applications~\cite{iacopini2018network}. Traditional methods, such as historical data analysis, often fall short in scalability and struggle to keep pace with rapid advancements. 

\noindent \textbf{The Interdisciplinary Future of Innovation}~
Interdisciplinary research, which often serves as the pivotal role for major breakthroughs, presents another significant challenge. With the rapid growth of scientific literature across diverse fields, manual identification of promising cross-disciplinary opportunities has become increasingly unfeasible~\cite{bolt2021educating}. The complexity and scale of this task call for automated solutions capable of discovering novel connections across fields.

\subsubsection{Advantages of AI4SoS}
AI offers an opportunity for tackling challenges in the SoS by leveraging its capacity to process vast datasets and identify complex patterns beyond human discernment. In the context of forecasting technological development, AI models can analyze citation networks, research metadata, and publication trends to detect emerging technological trajectories with enhanced precision~\cite{borner2018forecasting}.

Moreover, AI-driven methods excel in uncovering interdisciplinary opportunities by representing scientific knowledge as graph structures and employing advanced similarity metrics. Graph neural networks, for instance, have demonstrated the ability to model intricate relationships across scientific literature, facilitating the discovery of latent connections and novel collaborations across disparate domains~\cite{zhou2020graph}. This capability empowers researchers to target high-potential interdisciplinary collaborations, fostering innovation at the convergence of fields.

\subsection{Understanding the Dynamics of Research Society}
\subsubsection{Background of Problem}
The dynamics of research societies play a fundamental role in shaping scientific progress, which encompass how scientist research patterns evolve, how different team constructions influence the impact of research output, and how current research society influences scientists.

\noindent \textbf{The Dynamics and Mechanics of Scientist Career}~
The role of studying scientific careers is to provide personalized support to the academic community, thereby enhancing individual innovation capabilities, optimize team collaboration efficiency, and improving the allocation of research resources~\citep{fortunato2018science}. However, challenges include the highly individualized nature of career development paths, data scarcity and bias, and the complexity of external environmental factors~\citep{wang2019early}. 

\noindent \textbf{The Dynamics and Mechanics of Research Team}~
The composition and dynamics of scientific teams play a crucial role in improving research outcomes, with elements such as size, diversity, and collaboration patterns influencing team creativity and productivity~\cite{alshebli2018preeminence,wu2019large}. Over time, shifts in team structures and researcher mobility have reflected broader changes in the research landscape. Understanding these evolving dynamics presents challenges, as the relationships between team composition and research impact are multifaceted~\cite{wuchty2007increasing,yang2022gender}. 

\noindent \textbf{The Dynamics and Mechanics of Research Society}~
The organization and dynamics of research societies play a crucial role in shaping the progression and fairness of scientific endeavors. Studies have highlighted persistent inequalities in academic representation, participation, and recognition, both within and across nations~\cite{wapman2022quantifying,liu2023non}. These disparities, influenced by systemic and structural factors, hinder the equitable generation and dissemination of knowledge. On a broader scale, imbalances in citation patterns and collaboration networks often reflect biases rooted in reputation and resources rather than research quality~\cite{gomez2022leading}. 

\subsubsection{Advantages of AI4SoS}
AI offers potential for understanding and improving the dynamics of research societies. By analyzing large-scale historical datasets—such as collaboration patterns, research trajectories, and external influences—AI can uncover critical factors driving individual career development. This enables personalized researcher support and helps institutions optimize talent management. Techniques such as predictive modeling have proven effective in tracking and forecasting team member mobility patterns~\cite{guimera2005team}.

Moreover, AI-driven agents can simulate complex team dynamics, providing insights into how various factors, such as diversity and team size, influence research productivity and innovation. Taking this a step further, AI can simulate entire scientific societies, not only uncovering hidden patterns and problems but also guiding the policymaking process by validating potential policies within the simulated environment. For instance, multi-agent systems have been employed to model team formation processes and predict collaboration outcomes under varying settings~\cite{su2024two}.

\section{Challenges and Pathways} \label{sec:challenges}
Achieving fully automated SoS discovery centers on effectively utilizing AI techniques to process scientific data. This endeavor involves addressing four key challenges: data-related issues, comprehensive system construction, robust system evaluation, and system explainability. For each of these challenges, we provide a detailed analysis along with potential pathways for resolution.

\subsection{Data Issues}
\noindent \textbf{Challenges}~Data issues mainly include data imbalance across disciplines and training data bias.
For the first issue, many disciplines, such as computer science and engineering, produce large volumes of well-structured data readily used by AI systems~\cite{fernandez2017insight,kaur2019systematic}. However, other fields, such as social sciences or humanities, often suffer from smaller datasets, less structured data, or incomplete information, which makes it difficult for AI models to provide accurate predictions~\cite{leevy2018survey,johnson2019survey}. This imbalance can lead to skewed results where AI predictions are disproportionately driven by well-represented fields, neglecting potentially valuable insights from underrepresented areas of research.
Another issue is training data bias. When predicting reproducible patterns from data, machine learning models inevitably incorporate and perpetuate biases present in the data, often in opaque ways~\citep{liu2023data}. For example, the training data and alignment methods of LLMs (whether open-source or closed-source) are not fully disclosed~\citep{achiam2023gpt,dubey2024llama,yang2024qwen2}, making it impossible to objectively assess their bias and fairness. Therefore, the fairness of machine learning becomes a heavily debated issue in applications ranging from the criminal justice system to hiring processes~\citep{mehrabi2021survey}.

\begin{table*}[ht]
\centering
\begin{threeparttable}
\caption{Summary table of large-scale cross-discipline academic datasets.}
\label{tab:datasets}
\begin{tabularx}{\textwidth}{X>{\raggedright\arraybackslash}p{0.24\textwidth}>{\raggedright\arraybackslash}p{0.24\textwidth}>{\raggedright\arraybackslash}p{0.25\textwidth}}
\toprule
Datasets & MAG & OAG & SciSciNet  \\
\midrule
Due & 2020 & 2023 & 2021  \\
Domain & Art, Biology, Business, Chemistry, Computer Science, Economics, Engineering, Environmental Science, Geography, Geology, History, Materials Science, Mathematics, Philosophy, Physics, Political Science, Psychology, Sociology & Art, Biology, Business, Chemistry, Computer Science, Economics, Engineering, Environmental Science, Geography, Geology, History, Materials Science, Mathematics, Philosophy, Physics, Political Science, Psychology, Sociology & Art, Biology, Business, Chemistry, Computer Science, Economics, Engineering, Environmental Science, Geography, Geology, History, Materials Science, Mathematics, Medicine, Philosophy, Physics, Political Science, Psychology, Sociology \\ 
Author &  261,445,825 & \phantom{0}35,774,510 &  134,197,162 \\
Paper & 247,389,875 & 130,710,733 & 134,129,188 \\
Affiliation & \phantom{000,0}25,811 & \phantom{000,}143,749 & \phantom{000,0}26,998 \\
\bottomrule
\end{tabularx}
\end{threeparttable}
\end{table*}

\noindent \textbf{Pathway}~To address issues of data imbalance and biases in training data, constructing a large and diverse dataset is essential to improve data representativeness, ensuring coverage across various domains, groups, and contexts. Several large-scale, cross-disciplinary academic datasets are currently available for SoS research, including the Microsoft Academic Graph (MAG)~\cite{sinha2015overview}, Open Academic Graph (OAG)\citep{zhang2022oag}, and SciSciNet~\citep{lin2023sciscinet}, where the statistical information of each dataset is summarized in Table~\ref{tab:datasets}. In the process of data auditing and filtering, it is crucial to examine data sources and mitigate any potential historical or socio-cultural biases to ensure the dataset is free from implicit biases~\citep{scatiggio2020tackling}. Additionally, employing multi-annotator strategies, conducting group balance checks, and performing fairness evaluations can further ensure the fairness and diversity of the dataset~\citep{prabhakaran2021releasing}. These measures not only enhance the model’s generalization ability but also reduce unfairness stemming from data biases.

\subsection{Comprehensive System Construction}
\noindent \textbf{Challenges}~Simulating a research society using AI for fully automated SoS discovery, particularly through an agent-based system, presents numerous challenges. Each scientist-agent requires detailed modeling of their research expertise, career trajectory, and collaborative networks, which are often too complex to be fully captured in the simulation system~\cite{norling2017informal,gao2024large}. Critical but unobservable factors, such as internal cognitive processes and informal discussions that drive real-world decision-making, remain challenging to replicate accurately. These limitations inevitably make simulations discrete and less representative of actual societal dynamics. 
Moreover, the simulation process itself introduces complexities. Aligning the simulated timeline with real-world events necessitates careful calibration; for instance, determining how many simulation epochs correspond to a year in reality~\cite{khaleghian2023calibrating}. Determining the appropriate size of the simulated society is also crucial; an overly small-scale model risks failing to capture the emergent behaviors of a real research ecosystem, while an overly large model may become impractical to manage and analyze~\cite{schulze2017agent,an2021challenges}.
Another pressing challenge lies in bias amplification when designing AI systems—a concern that builds on the broader implications of how AI interacts with societal structures. Since AI systems are often designed to optimize based on historical data of SoS, they risk perpetuating existing paradigms, funding trends, and citation networks. This aligns with the well-documented “rich get richer” effect in citation and funding dynamics~\cite{ebadi2015receive,ronda2018evolutions,katz2020metrics}. If an AI system prioritizes high-impact metrics, it may inadvertently favor mainstream topics and established researchers, further marginalizing unconventional or disruptive ideas. Without explicit mechanisms to value novelty and diversity, such systems could unintentionally confine the scientific community to existing trends, hindering pathways to groundbreaking innovation.
Lastly, the system must account for unexpected exceptions to ensure the simulation operates smoothly and continuously for fully automated scientific discovery. Striking a balance between realism and feasibility remains a persistent and fundamental challenge in these simulations.

\noindent \textbf{Pathway}~Several potential pathways can help address these complexities.
With the continuous advancement of LLMs' comprehensive capabilities, handling complex multi-level modeling is becoming increasingly feasible. By defining agent models with distinct roles and appropriately assigning tasks, the behaviors of scientists at various levels can be more accurately simulated~\citep{qian2024chat}. Fine-tuning LLMs on extensive academic datasets can further optimize the behavioral patterns of agents~\citep{guo2024large}, enhancing their adaptability to reflect real-world research dynamics.
One solution for timeline alignment is to build flexible, dynamic calibration techniques that adjust the simulation's temporal parameters based on context and event-driven data~\citep{yang2024oasis}.
In determining the appropriate scale for the simulated society, agent-based sampling methods (random or rule-based) or dynamic population expansion techniques can be utilized~\citep{su2024two}.
When addressing bias in AI systems, it is crucial to consider the nature of SoS, a discipline dedicated to analyzing historical data and uncovering biases or patterns within the scientific community. To ensure alignment between simulations and real-world dynamics, it is essential to incorporate these biases into SoS studies, as AI designed for this field seeks to enhance and advance SoS research. At the same time, such biases can be mitigated through targeted adjustments to system parameters. For instance, to counteract the “rich get richer” effect in citations, one effective approach could involve reducing the likelihood of citing highly cited papers when an agent selects a reference. Instead, assigning higher probabilities to less-cited, more novel papers can help promote diversity in citation practices and encourage the exploration of unconventional ideas.
Moreover, the system can integrate robust anomaly detection and recovery mechanisms to handle unexpected situations. Using unsupervised learning techniques (such as clustering), the model can identify deviations from expected behaviors and adjust simulation parameters accordingly to ensure stability and continuity~\citep{aldrich2013unsupervised}.
These potential solutions try to strike a balance between realism and operational feasibility, providing a technological foundation for research society simulations.

\subsection{Comprehensive System Evaluation}
\noindent \textbf{Challenges}~Evaluating the validity of outputs generated by AI systems in the field of SoS is a complex and multifaceted challenge. SoS research addresses a broad range of problems and lacks unified evaluation standards, with different tasks often necessitating tailored metrics~\citep{liu2023data}. Moreover, innovation—a key attribute of AI outputs—is inherently subjective and context-dependent, making it difficult to quantify accurately using traditional methods~\citep{su2024two,chen2024philosopher}. Validity assessments also heavily rely on specific domain contexts. However, the interdisciplinary nature of SoS compounds the complexity, requiring the integration of knowledge and evaluation standards from diverse fields.
Additionally, the dynamic nature and long-term implications of AI-generated outputs present further challenges, as their true impact on scientific progress often cannot be evaluated in the short term~\citep{balasubramaniam2024road}. Addressing this requires advanced tools, such as time-series analysis and virtual scientist simulations, to facilitate longitudinal tracking. Furthermore, AI-generated scientific recommendations may raise ethical issues and have far-reaching consequences for scientific communities and research practices~\citep{lissack2024navigating}. Therefore, a comprehensive and adaptable evaluation framework is necessary, integrating scientometric methodologies, multidisciplinary expert reviews, dynamic analytical approaches, and stringent ethical guidelines.

\noindent \textbf{Pathway}~To address these challenges, appropriate solutions can be implemented. First, collaborating with domain experts to define task-specific evaluation metrics is essential, and then quantitative evaluation methods based on scientometrics should be developed. For instance, citation counts can be used as a measure of influence when evaluating the impact of system outputs, and they can also track knowledge flow~\citep{liu2023data}. In simulating a scientist's career, individual impact metrics such as the h-index, which reflects both productivity and impact, can be applied. Additionally, to assess output novelty, feasible approaches include large model-based peer-review scoring~\citep{lu2024ai,su2024two} or calculating the Z-score for each pairing of referenced journals~\citep{chen2024philosopher}.
With the ongoing expansion of LLMs' expertise and improved reasoning capabilities, interdisciplinary testing and long-term large-scale simulations have become increasingly feasible. Moreover, LLMs are now being employed in social simulations~\citep{yang2024oasis}, assuming role-based agents. In terms of ethical and social impacts, aligning model preferences and improving transparency can partially address ethical concerns and enhance user trust, while ethical benchmarks~\citep{meadows2024localvaluebench,ji2024moralbench} can be used to test the validity of system outputs. By integrating these strategies, a multidimensional evaluation framework can be established.

\subsection{Explainability and Causal Inference}

\noindent \textbf{Challenges}~While the AI framework emphasizes automated discovery and evaluation, it lacks mechanisms to explain the causal pathways behind AI-generated outputs~\cite{hassija2024interpreting,reddy2025towards}. This limitation makes it difficult for researchers and policymakers to trust and adopt AI-driven insights, as they may not fully understand the underlying logic or relationships. Moreover, the complex and interdisciplinary nature of SoS often involves interactions between numerous variables, such as collaborations, funding patterns, and citation networks~\cite{sonnenwald2007scientific,fortunato2018science}, which cannot be adequately captured through correlation-based approaches. Without explicit causal explanations, it is challenging to ensure the auditability, accountability, and interpretability of the system, undermining its credibility and ethical alignment.

\noindent \textbf{Pathway}~To address these challenges, it is crucial to introduce causal modeling~\cite{petersen2014causal,feuerriegel2024causal} and explainable AI (XAI)~\cite{dwivedi2023explainable,longo2024explainable} techniques to assist in interpreting and validating simulation results. Approaches such as Counterfactual Analysis can clarify the logical origins of AI-driven recommendations or discoveries, making the reasoning process more transparent. Relevant methods in the SoS domain include causal inference techniques like Propensity Score Matching (PSM) and Coarsened Exact Matching (CEM), which are useful for identifying causal relationships in complex systems~\cite{king2011comparative,imbens2015causal}. Additionally, causal graphical models and structural equation modeling (SEM) can be applied to analyze scientific impact by modeling the flow of influence across variables such as collaboration networks or funding distributions~\cite{de2014applications,khan2019methodological,leist2022mapping}. These tools provide a robust foundation for explaining AI-generated outputs.

\begin{figure}[t]
    \centering
    \includegraphics[width=\linewidth]{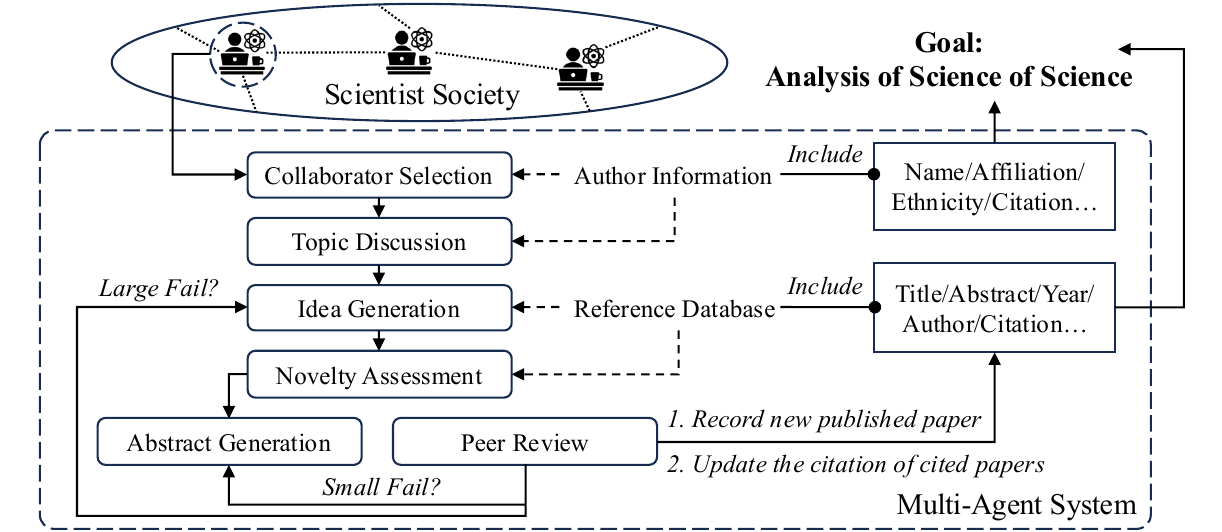}
    \caption{The overview of our preliminary multi-agent system for scientific collaboration simulation. We place the simulation within a community of scientists. After a scientist leads his/her team in submitting a paper, it undergoes peer review. If accepted, it is added to the reference database and can be cited by other scientists in subsequent epochs. Due to varying author information, the citation count of the final research output differs, then we can analyze the correlation between them—understanding the dynamics of research organizations, which is important in the field of SoS.}
    \label{fig:system_overview}
\end{figure}

\section{Proof-of-Concept Studies} \label{sec:experiment}

In this section, we present case studies to illustrate a practical application scenarios in AI4SoS. Specifically, by constructing a simplified preliminary multi-agent system to replicate phenomena observed in real-world scientific societies and uncover underlying patterns in SoS, we aim to demonstrate the possibility of automated pattern discovery.

\subsection{Environment Construction}
We construct a preliminary multi-agent system to simulate a society-level scientific collaboration through an end-to-end pipeline, including collaborator selection, topic discussion, idea generation, novelty assessment, abstract generation, and peer review, inspired by~\citep{lu2024ai,qi2024large,su2024two}. Existing studies primarily focus on simulating individual scientists or small research teams within specific fields (e.g., computer science) and are often constrained to isolated settings that do not capture the broader research ecosystem. In contrast, our work enhances the system's complexity by incorporating realistic factors such as multidisciplinary data, a review and indexing system, and scalable simulation across multiple research teams. The overview of our system is shown in Fig.~\ref{fig:system_overview}.

\begin{table*}[t]
\centering
\begin{threeparttable}
\caption{Summary table of disciplines and fields~\citep{alshebli2018preeminence}.}
\label{tab:discipline}
\begin{tabularx}{\textwidth}{>{\raggedright\arraybackslash}p{0.4\textwidth}X}
\toprule
Field & Discipline \\
\midrule
Humanities, Literature \& Arts & [Art, History, Philosophy, Psychology] \\
Life Science \& Earth Sciences & [Biology, Environmental Science, Geography, Geology]  \\
Business, Economics \& Management & [Business, Economics]  \\
Engineering \& Computer Science & [Computer Science, Engineering]\\
Chemical \& Material Sciences & [Chemistry, Materials Science]  \\
Physics \& Mathematics & [Mathematics, Physics]\\
Health \& Medical Sciences & [Medicine]\\
Social Sciences & [Political Science, Sociology]\\
\bottomrule
\end{tabularx}
\end{threeparttable}
\end{table*}

\begin{table*}[t]
\centering
\begin{threeparttable}
\caption{Different strategies are adopted for various pieces of information regarding authors.}
\label{tab:information_1}
\begin{tabularx}{\textwidth}{>{\raggedright\arraybackslash}p{0.14\textwidth}X>{\raggedright\arraybackslash}p{0.22\textwidth}}
\toprule
Field Name & Strategy & Example \\
\midrule
\multicolumn{3}{c}{\textit{Author Information}}\\
Name & Use the anonymization technique & Scientist 1 \\
Ethnicity & Use the name ethnicity classifier~\citep{ambekar2009name} & British \\
Affiliation & Retain the original content & [King’s College London] \\
Affiliation Ranking & Use THE World University Rankings 2025~\tnote{1} & 36\\
Citation & Extract the author's published papers between 2010 to 2020 and calculate the total number of citations for the papers; In the simulation, it will be updated if his/her paper is cited & 1800 \\
Co-author & Extract the author's published papers between 2010 to 2020 and record the collaborators in the papers; In the simulation, it will be updated if there are new collaborators & [Scientist 10, Scientist 201, Scientist 1002, \dots] \\
Discipline & Extract the author's published papers between 2010 to 2020 and assign the author's discipline as the one that appears most frequently  & Psychology\\
Research topic & Extract the author's published papers between 2010 to 2020 and record the keywords in the papers; Use GPT-4 to summarize these keywords into research topics & [Neuropsychology, Cognitive flexibility, Attentional bias, \dots] \\
\bottomrule
\end{tabularx}
\begin{tablenotes}
\item[1] \url{https://www.timeshighereducation.com/world-university-rankings/latest/world-ranking}
\end{tablenotes}
\end{threeparttable}
\end{table*}

\noindent \textbf{Multidisciplinary Data}~
We use the OAG 3.1\footnote{\url{https://open.aminer.cn/open/article?id=65bf053091c938e5025a31e2}} as the initial database for our system, which developed from the Open Academic Graph~\citep{zhang2022oag}. This data set includes 35,774,510 authors and 130,710,733 papers as of 2023, spanning diverse domains such as physics, chemistry, and computer science. In Table~\ref{tab:discipline}, we present the disciplines and fields of paper in the Open Academic Graph, which is used to analyze the potential different patterns in various areas. We use papers from 2002 to 2009 as the reference database and papers from 2010 to 2011 as the validation database. To address missing author ethnicity and paper field information—key elements for validating SoS findings—we employ several data completion strategies. Specifically, we adopt corresponding approaches for the various pieces of author information and paper information in this dataset for our simulation, shown in Table~\ref{tab:information_1} and~\ref{tab:information_2}.

\begin{table*}[ht]
\centering
\begin{threeparttable}
\caption{Different strategies are adopted for various pieces of information regarding papers.}
\label{tab:information_2}
\begin{tabularx}{\textwidth}{>{\raggedright\arraybackslash}p{0.14\textwidth}X>{\raggedright\arraybackslash}p{0.22\textwidth}}
\toprule
Field Name & Strategy & Example \\
\midrule
\multicolumn{3}{c}{\textit{Paper Information}}\\
Title & Retain the original content & Linkages of plant traits to soil properties \dots\\
Abstract & Retain the original content & Global change is likely to alter plant community \dots\\
Year & The year of the papers in the initial database is set to -1, while the papers published by the agent are assigned the epoch when the review is accepted & -1 \\
Citation & In the initial database, the citation count of the papers is the original citation value plus the number of times they are cited during the simulation, while the citation count of the papers written by the agent is the number of times they are cited during the simulation & 82 \\
Authors & Retain the original content & [Scientist 124, Scientist 7923, \dots]\\
Cited Paper & The papers in the initial database have None for this information due to its absence, while the papers published by the agent contain the names of the cited papers & None \\
Discipline & Use GPT-4 to classify the papers into disciplines based on their keywords and titles. Refer to Table~\ref{tab:discipline} for all the disciplines used & Environmental Science\\
\bottomrule
\end{tabularx}
\end{threeparttable}
\end{table*}

\noindent \textbf{Review and Indexing System}~
To better simulate and reveal the patterns of scientific collaboration mechanisms, we introduce a review and indexing system. Papers written by scientist teams are peer-reviewed and scored (ranging from 1 to 10), and those that exceed the acceptance threshold (with score larger than 5) are added to the reference paper database as newly published papers. 
The peer review criteria are discussed in Appx.~\ref{sec:review}, considering that the outcomes are cross-disciplinary.
Besides, the indexing system allows agents to retrieve published papers as references, and the citation count of referenced papers is updated accordingly, which is later used for metric evaluation.

\begin{figure}[t]
    \centering
    \includegraphics[width=0.8\linewidth]{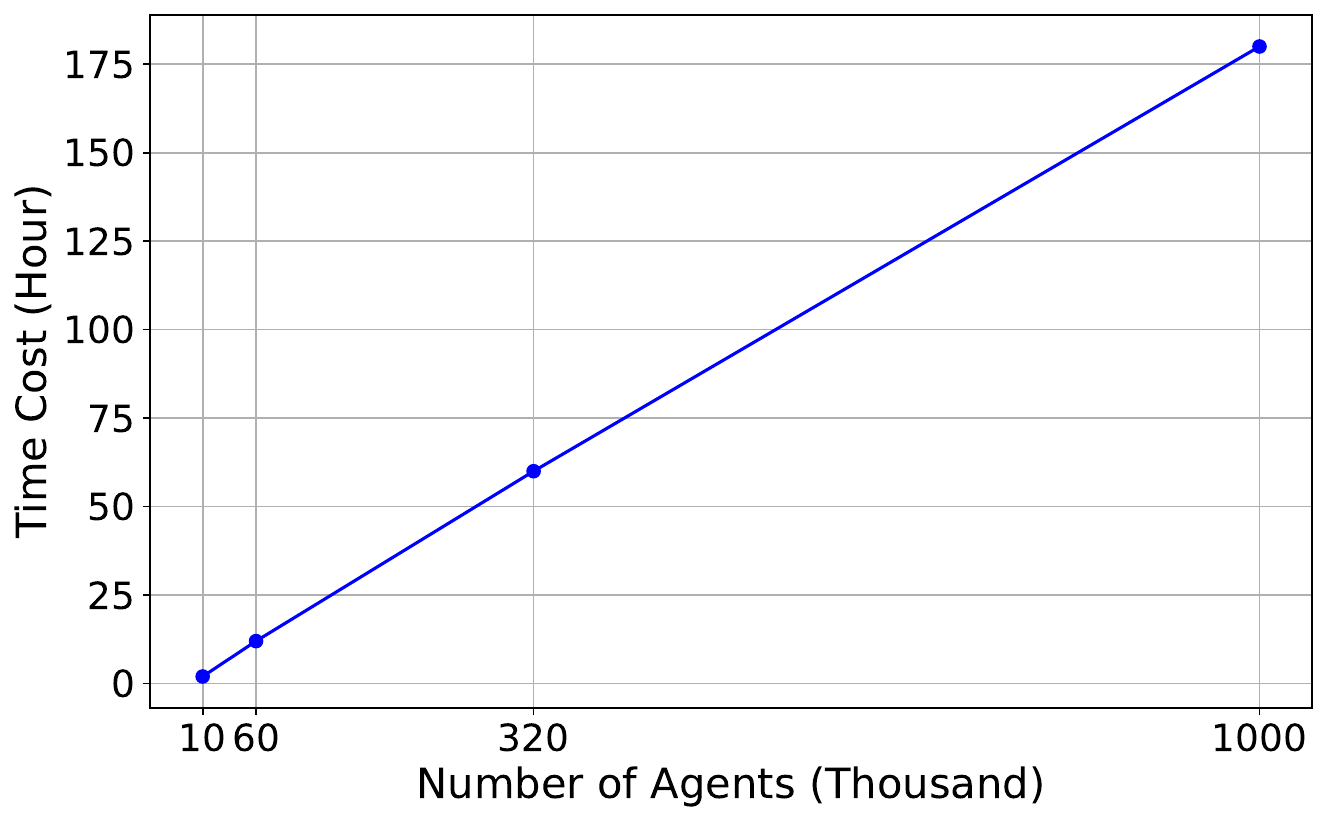}
    \caption{The time taken for a complete scientific collaboration with agents of different scales. A simulation of a million-agent society takes only one week.}
    \label{fig:sample_image}
\end{figure}

\noindent \textbf{Scalable Simulation}~
To better replicate the phenomenon of free collaboration in real scientific cooperation, we implement an adaptive concurrent distributed system based on the OASIS~\citep{yang2024oasis}. The system's asynchronous mechanism achieves concurrent processing by queuing multiple requests from agents in an inference channel and then distributing them to different ports for sending and receiving, where each port has deployed an LLM responsible for chatting or embedding. Furthermore, to reduce CPU load, we set the channel allocation wait time based on the number of pending requests in the channel, thereby enabling long-term large-scale asynchronous simulation. This mechanism serves the two purposes: 1. Enabling scientist agents from different teams to communicate simultaneously, including both intra-team and cross-team collaboration, and 2. Accelerating the simulation process to enable large-scale simulations at the million-agent level.
We test the time cost of our simulation system under different number of agents, illustrated in Fig.~\ref{fig:sample_image}. It could be found that we realize a fast large-scale agent system, where a simulation of a million agent society takes only one week. 

\begin{figure}[t]
    \centering
    \includegraphics[width=0.8\linewidth]{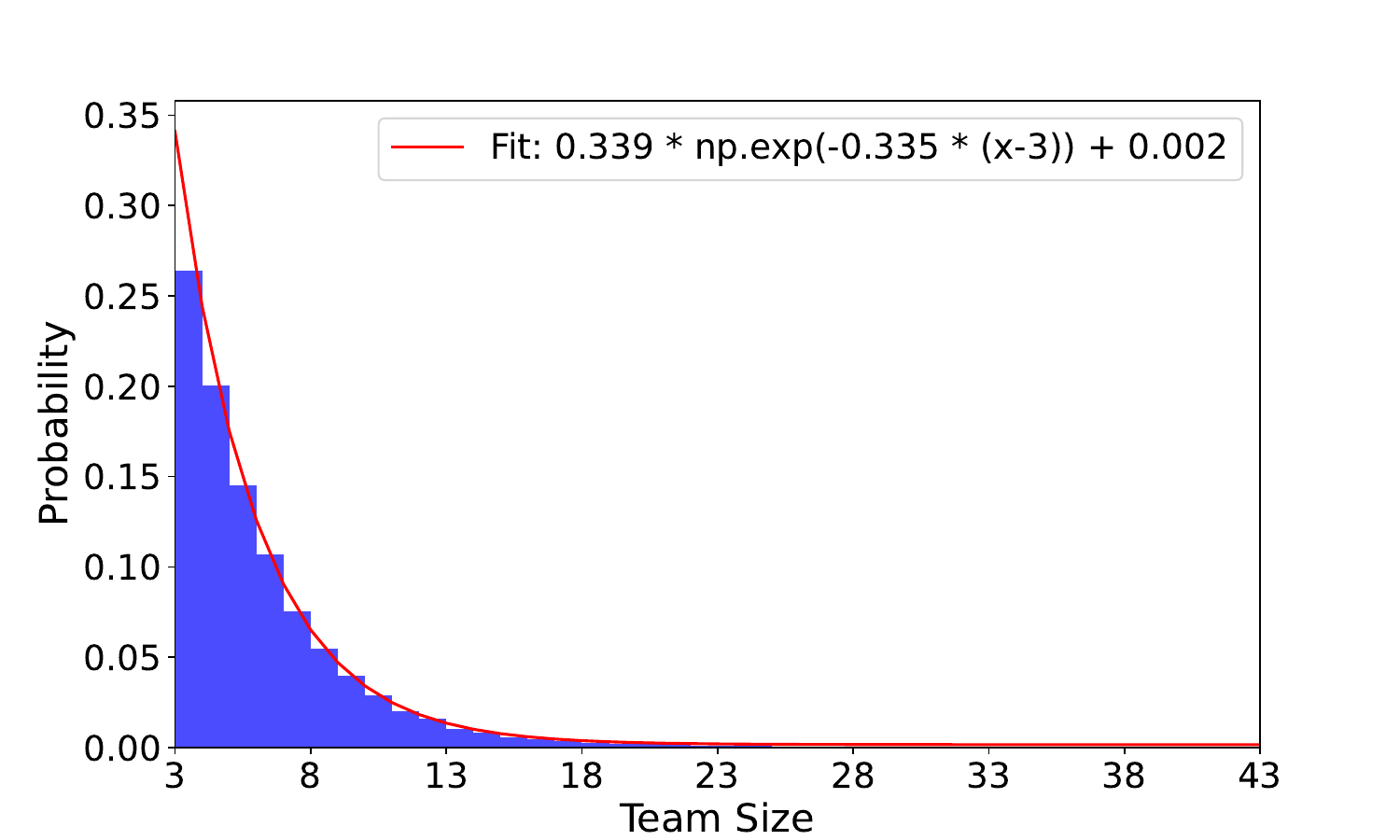}
    \caption{The statistics of team sizes for papers published between 2002 and 2009 in the OAG, with the red fitting line revealing that the distribution follows an exponential pattern.}
    \label{fig:team_size}
\end{figure}

\subsection{Experiments}
\noindent \textbf{Implementation Details}
We implement our system on 32 NVIDIA A100 GPUs, with 4 ports deployed on each GPU, and each port running the \textit{LLaMA3.1-8b} model. 
We allow each agent to create up to 3 teams simultaneously, with team sizes following an exponential distribution. This is because we analyze the team sizes of papers published between 2002 and 2009 in the OAG (over 1,000,000 papers), as shown in Fig.~\ref{fig:team_size}. The red fitting line indicates that the team sizes in the real data follow an exponential distribution. Therefore, in our simulation, the team size of each agent is also modeled using an exponential distribution.

In idea generation and novelty assessment, each agent can cite up to 9 references per speech, where the retrieval results are obtained based on the similarity between the embeddings of the query terms and the embeddings of the papers in the database. The model used for embedding is \textit{mxbai-embed-large}. To avoid storage issues, each agent's memory retains a maximum of 5 entries. Each paper undergoes peer review by 3 reviewers. In terms of the timeline, each epoch allows for 1 action, meaning a complete scientific collaboration can be completed in 6 epochs if the team progresses without any delays or interruptions. In our final experiment, the size of our society is maintained at 1 million agents, with a total of 40 epochs.

\begin{figure*}[t]
    \centering
    \includegraphics[width=\linewidth]{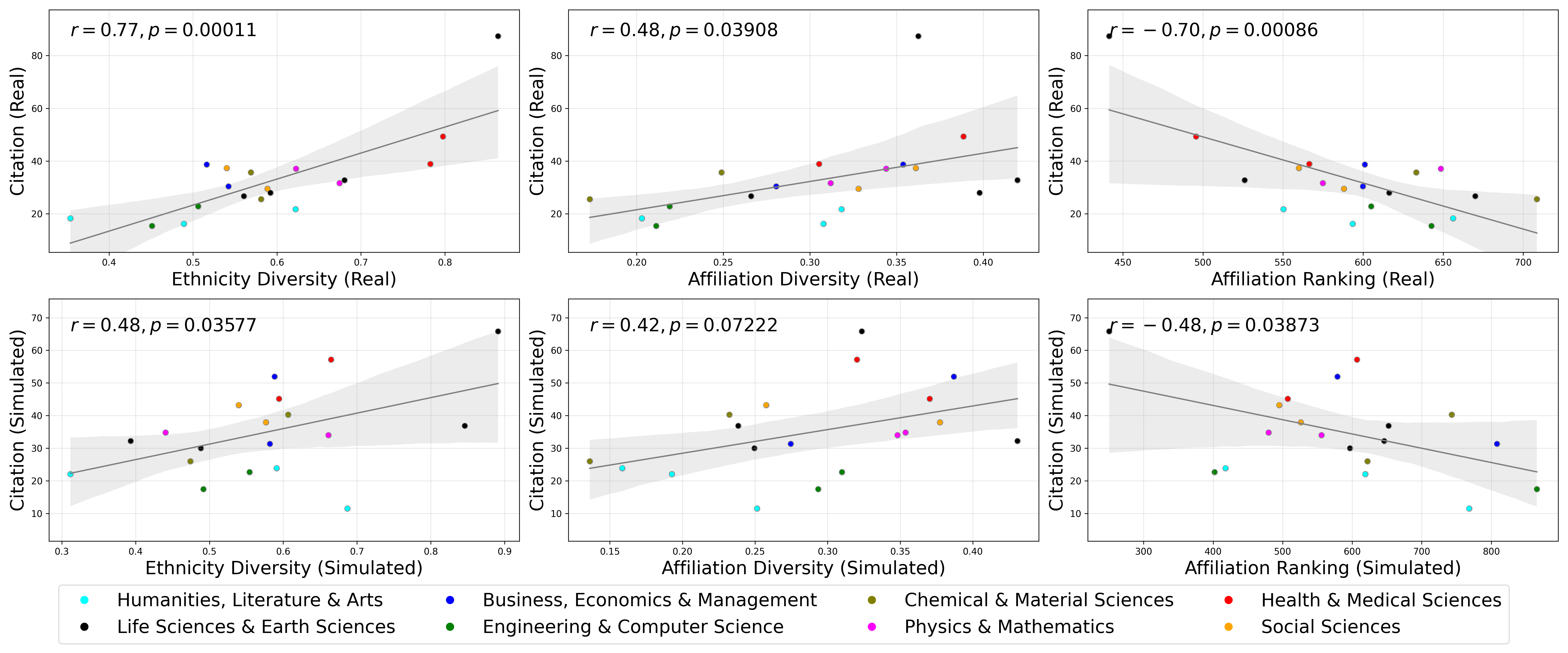}
    \caption{Comparison of real-world (2010) and AI-simulated scientific research patterns. The scatter plots illustrate the relationships between Ethnicity Diversity, Affiliation Diversity, and Affiliation Ranking with Citation Count in both real-world (top row) and simulated (bottom row) data. Strong correlations observed in real data are partially reproduced by the AI-driven multi-agent system, demonstrating its potential to uncover meaningful patterns in scientific research and support automated SoS studies.}
    \label{fig:experimental_results}
\end{figure*}

\noindent \textbf{Involved Metrics}~
Following the settings of~\cite{alshebli2018preeminence,li2019early, wang2019early}, we measure the impact of scientific output by the number of citations a paper receives. In the simulation, the citation counts are updated each time a paper is retrieved during the idea generation phase. For validation, we analyze the citation counts of agent-generated papers to assess whether the system can replicate patterns observed in real-world data from the years 2010 to 2011.
To evaluate AI’s potential in pattern discovery, we examine the influence of three key factors on citation counts: ethnicity diversity, affiliation diversity, and average university ranking. Specifically, we measure diversity using Shannon entropy. For instance, the ethnicity diversity $d_{eth}$ of paper $s$ is calculated as:
\begin{gather}
d_{eth} = -\sum_{i=1}^{k}p_{i}(s)\ln{p_i(s)},
\end{gather}
where $k$ represents the total number of ethnicity categories, and $p_i(s)$ is the proportion of authors from the $i$-th ethnicity category in paper $s$.

\begin{figure*}[ht]
    \centering
    \includegraphics[width=\linewidth]{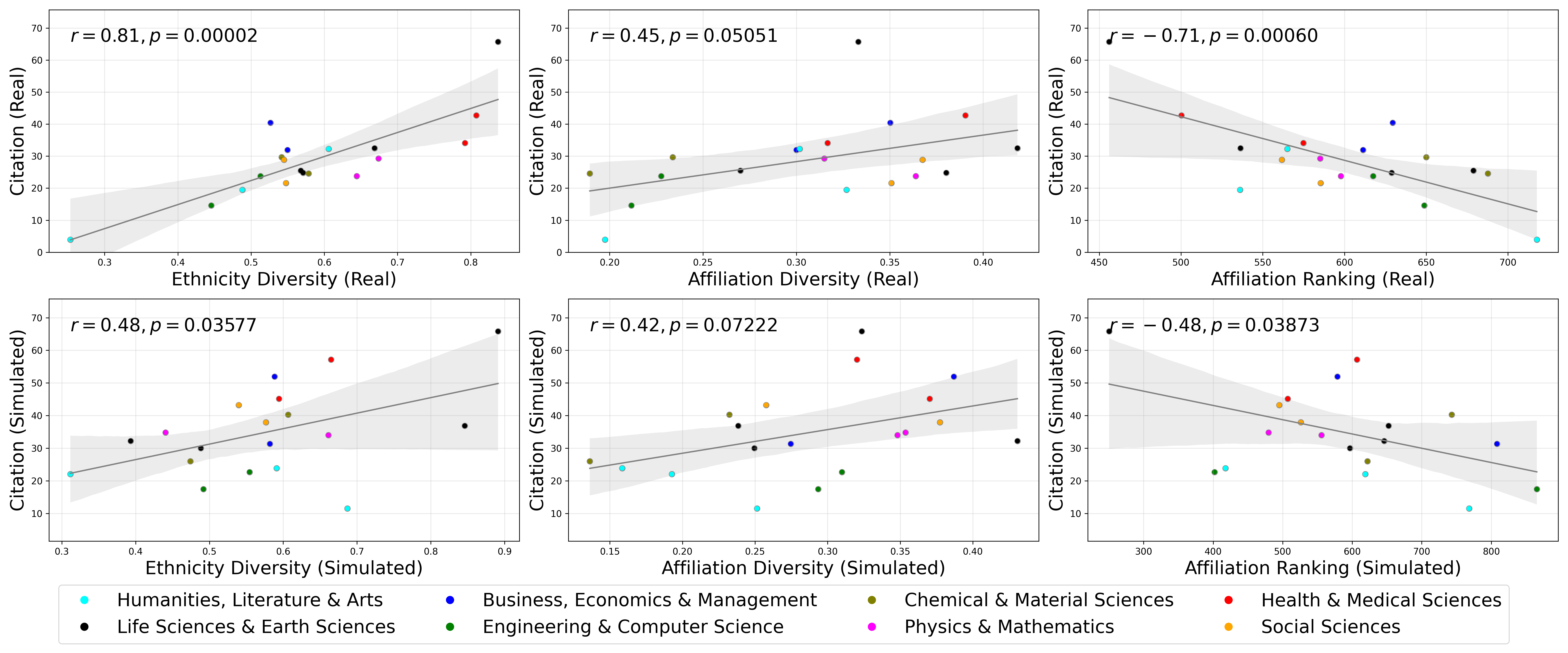}
    \caption{Comparison of real-world (2011) and AI-simulated scientific research patterns.}
    \label{fig:experimental_results_2011}
\end{figure*}

\subsection{Simulation Results}
The experimental results presented in Fig.~\ref{fig:experimental_results} compare real-world data in 2010 with the outcomes generated by our preliminary LLM-based multi-agent system. Both the real-world and simulated data show that higher citation counts are positively correlated with greater ethnicity diversity, which aligns with existing findings in SoS literature~\cite{alshebli2018preeminence}, although the correlations are slightly weaker in the simulation. Additionally, the negative correlation between affiliation ranking and citation counts is also reproduced in the simulated data, suggesting that institutions with higher rankings may achieve higher citation counts per research output.

A similar comparison using real-world data from 2011 and the simulated result is provided in Fig.~\ref{fig:experimental_results_2011}. The statistical analysis of the 2011 data exhibits similar trends to those observed in Fig.~\ref{fig:experimental_results}, which presents the comparison using 2010 data. The positive correlation between citation counts and ethnicity diversity, as well as the negative correlation between affiliation ranking and citation counts, are consistently reflected in both years. However, minor variations in correlation strength are observed, highlighting the dynamic nature of scientific collaboration trends over time.

However, while both real-world and simulated data indicate a positive correlation between citation counts and affiliation diversity, the pattern observed in the simulation is not statistically significant, with a p-value greater than 0.05. These results suggest that the preliminary AI-driven simulations have the potential to replicate and uncover key patterns in scientific research, but there remains significant room for improvement.
For instance, the current system lacks several critical components, such as comprehensive modeling of individual research trajectories and realistic funding and policy influences. These limitations contribute to the preliminary nature of our approach, as the absence of such factors restricts the system's ability to fully capture the complexity of real-world scientific ecosystems. Developing a more comprehensive and sophisticated simulation framework will enhance the system’s capability to automatically model complex scientific dynamics with greater accuracy and reliability.

\section{Alternative Views}
The application of AI in SoS is often seen as transformative, promising to accelerate discovery. However, critics highlight significant limitations and risks, questioning its unqualified benefits. These concerns focus on systemic issues and unintended consequences~\citep{binns2018fairness,raji2019actionable,messeri2024artificial}. Key counterarguments include:
(1) Reinforcement of Existing Inequalities:
AI systems rely heavily on historical data, which often mirror long-standing inequities within the scientific community. For instance, datasets may disproportionately represent well-established disciplines, regions, or researchers, thereby perpetuating an imbalanced view of scientific contributions. Critics argue that this could stifle innovation by overlooking emerging fields and underrepresented groups, ultimately reinforcing the leading trend rather than fostering diversity. (2) Overreliance on Traditional Metrics:
Academic evaluation metrics, such as citation counts and journal impact factors, are central to many AI applications in SoS. These metrics have been criticized for prioritizing mainstream research while marginalizing unconventional or nascent ideas. Opponents caution that AI-driven analyses might amplify this bias, narrowing the scope of scientific discovery and undervaluing novel contributions.


While these critiques highlight significant challenges, they underscore the importance of addressing fairness, and inclusivity in AI applications for SoS~\citep{holstein2019improving,schwartz2021proposal,schwartz2022towards}. To mitigate these concerns, the following strategies can be adopted:
(1) Promoting Diversity in Data and Metrics:
Expanding data curation efforts to include a wider range of disciplines, regions, and research communities is critical for minimizing biases. Additionally, developing diversified scientific impact metrics beyond citation counts can ensure a more equitable evaluation of research contributions.
(2) Incorporating Bias Mitigation Techniques:
Embedding bias detection and correction mechanisms in AI systems can help identify and address inequities in the data and algorithms. These techniques should be complemented by rigorous validation to ensure fairness and reliability.

\section{Outlook}
As AI4SoS progresses toward full autonomy, we envision a future where scientific discovery itself becomes a more self-reflective, adaptive, and strategically guided process. In this envisioned landscape, AI agents are trained on vast corpora of scholarly data and historical innovation patterns, which will not only map the contours of scientific fields but also anticipate emerging disciplines and recommend actionable research agendas.

Automated SoS systems will continuously monitor the evolving structure of scientific collaboration, offering dynamic guidance to policymakers, institutions, and individual researchers. Research teams may be formed or optimized based on predicted synergy and complementary expertise, while funding strategies could adapt in real time to maximize long-term innovation impact.
Moreover, AI4SoS could democratize scientific foresight, making sophisticated analyses accessible to a broader range of stakeholders, from early-career researchers to global research organizations. The resulting ecosystem would be one where science is not only accelerated but also made more transparent, inclusive, and responsive to societal needs.

{To enhance real-world applicability, we also envision deployment scenarios in which AI4SoS integrates directly with existing scientific ecosystems. For instance, it could serve as a sandbox environment for evaluating national research policies, allowing simulated assessments before implementation. Within academic institutions, AI4SoS could support internal research strategy formulation, identifying growth areas and optimizing resource allocation. Additionally, it could assist governmental and funding bodies in planning emerging discipline layouts and national innovation agendas. These integration pathways would significantly boost the practical value, societal impact, and credibility of AI4SoS.}

Achieving this vision will demand sustained interdisciplinary collaboration, ethical oversight, and robust infrastructure, but the potential payoff is immense: a future in which the SoS is not just studied, but actively shaped by intelligent systems.

\section{Conclusion}
This paper presents a forward-looking perspective on the future of AI4SoS, proposing a five-level autonomy framework for understanding the progression toward automated SoS discovery.
We emphasize the importance of AI4SoS by demonstrating its potential in two critical domains: forecasting trends in technology and innovation, and analyzing the evolution of research communities. Furthermore, we discuss key challenges and future directions, supporting our vision with literature reviews and proof-of-concept studies that showcase early applications.
Ultimately, AI4SoS holds the promise of enabling automated SoS discovery, thereby enhancing scientific efficiency and promoting interdisciplinary innovation.



\section*{Impact Statement}

We believe that sustained collaboration between AI researchers and SoS scholars is essential for advancing our understanding of complex scientific processes. This study leverages the complementary expertise of both fields to address key SoS challenges, improving scientific efficiency and fostering interdisciplinary innovation.

However, from an ethical perspective, the integration of AI with SoS research may present several concerns. First, \textbf{accountability:} When AI participates in scientific decision-making, it is crucial to clarify responsibility. For instance, if an AI-generated prediction leads to errors, should developers bear full responsibility? We suggest enhancing AI system transparency (e.g., recording decision-making pathways) and explainability (e.g., providing reasoning behind decisions) to help researchers and regulators delineate accountability more clearly. Second, \textbf{fairness and bias:} AI systems rely on training data, which may contain inherent biases related to gender, geography, or economic disparities. These biases can lead to unjust scientific conclusions. Therefore, AI development and application should include rigorous data preprocessing and incorporate fairness constraints within algorithms to mitigate the risk of bias propagation. Finally, \textbf{public trust:} AI-driven automation tools, due to their complexity, may create a sense of detachment among the public. When AI decision-making processes are opaque, concerns about the credibility of scientific findings may arise. To foster trust, it is essential to develop more interpretable AI models and ensure human oversight in scientific processes.

From a societal perspective, the complexity of SoS demands innovative approaches. Conventional statistical studies, which depend largely on historical data, frequently struggle to uncover causal mechanisms. In contrast, agent-based AI provides a dynamic, causality-driven alternative. By elucidating the mechanisms behind the evolution of scientific knowledge, these methods can clarify how government policies influence research funding, academic publishing, and interdisciplinary collaboration. As AI4SoS advances, it will foster more effective knowledge exchange among academia, industry, and government, accelerating technological and theoretical innovation. Through intelligent analysis and predictive modeling, researchers can more precisely identify scientific challenges, significantly enhancing the efficiency of discovery.




\section*{Acknowledgements}
This work is supported by Shanghai Artificial Intelligence Laboratory. 

\bibliography{ref}


\begin{thebibliography}{110}
\ifx \bisbn   \undefined \def \bisbn  #1{ISBN #1}\fi
\ifx \binits  \undefined \def \binits#1{#1}\fi
\ifx \bauthor  \undefined \def \bauthor#1{#1}\fi
\ifx \batitle  \undefined \def \batitle#1{#1}\fi
\ifx \bjtitle  \undefined \def \bjtitle#1{#1}\fi
\ifx \bvolume  \undefined \def \bvolume#1{\textbf{#1}}\fi
\ifx \byear  \undefined \def \byear#1{#1}\fi
\ifx \bissue  \undefined \def \bissue#1{#1}\fi
\ifx \bfpage  \undefined \def \bfpage#1{#1}\fi
\ifx \blpage  \undefined \def \blpage #1{#1}\fi
\ifx \burl  \undefined \def \burl#1{\textsf{#1}}\fi
\ifx \doiurl  \undefined \def \doiurl#1{\url{https://doi.org/#1}}\fi
\ifx \betal  \undefined \def \betal{\textit{et al.}}\fi
\ifx \binstitute  \undefined \def \binstitute#1{#1}\fi
\ifx \binstitutionaled  \undefined \def \binstitutionaled#1{#1}\fi
\ifx \bctitle  \undefined \def \bctitle#1{#1}\fi
\ifx \beditor  \undefined \def \beditor#1{#1}\fi
\ifx \bpublisher  \undefined \def \bpublisher#1{#1}\fi
\ifx \bbtitle  \undefined \def \bbtitle#1{#1}\fi
\ifx \bedition  \undefined \def \bedition#1{#1}\fi
\ifx \bseriesno  \undefined \def \bseriesno#1{#1}\fi
\ifx \blocation  \undefined \def \blocation#1{#1}\fi
\ifx \bsertitle  \undefined \def \bsertitle#1{#1}\fi
\ifx \bsnm \undefined \def \bsnm#1{#1}\fi
\ifx \bsuffix \undefined \def \bsuffix#1{#1}\fi
\ifx \bparticle \undefined \def \bparticle#1{#1}\fi
\ifx \barticle \undefined \def \barticle#1{#1}\fi
\bibcommenthead
\ifx \bconfdate \undefined \def \bconfdate #1{#1}\fi
\ifx \botherref \undefined \def \botherref #1{#1}\fi
\ifx \url \undefined \def \url#1{\textsf{#1}}\fi
\ifx \bchapter \undefined \def \bchapter#1{#1}\fi
\ifx \bbook \undefined \def \bbook#1{#1}\fi
\ifx \bcomment \undefined \def \bcomment#1{#1}\fi
\ifx \oauthor \undefined \def \oauthor#1{#1}\fi
\ifx \citeauthoryear \undefined \def \citeauthoryear#1{#1}\fi
\ifx \endbibitem  \undefined \def \endbibitem {}\fi
\ifx \bconflocation  \undefined \def \bconflocation#1{#1}\fi
\ifx \arxivurl  \undefined \def \arxivurl#1{\textsf{#1}}\fi
\csname PreBibitemsHook\endcsname

\bibitem{fortunato2018science}
\begin{barticle}
\bauthor{\bsnm{Fortunato}, \binits{S.}},
\bauthor{\bsnm{Bergstrom}, \binits{C.T.}},
\bauthor{\bsnm{B{\"o}rner}, \binits{K.}},
\bauthor{\bsnm{Evans}, \binits{J.A.}},
\bauthor{\bsnm{Helbing}, \binits{D.}},
\bauthor{\bsnm{Milojevi{\'c}}, \binits{S.}},
\bauthor{\bsnm{Petersen}, \binits{A.M.}},
\bauthor{\bsnm{Radicchi}, \binits{F.}},
\bauthor{\bsnm{Sinatra}, \binits{R.}},
\bauthor{\bsnm{Uzzi}, \binits{B.}}, \betal:
\batitle{Science of science}.
\bjtitle{Science}
\bvolume{359}(\bissue{6379}),
\bfpage{0185}
(\byear{2018})
\end{barticle}
\endbibitem

\bibitem{bettencourt2009scientific}
\begin{barticle}
\bauthor{\bsnm{Bettencourt}, \binits{L.M.}},
\bauthor{\bsnm{Kaiser}, \binits{D.I.}},
\bauthor{\bsnm{Kaur}, \binits{J.}}:
\batitle{Scientific discovery and topological transitions in collaboration networks}.
\bjtitle{Journal of Informetrics}
\bvolume{3}(\bissue{3}),
\bfpage{210}--\blpage{221}
(\byear{2009})
\end{barticle}
\endbibitem

\bibitem{shi2015weaving}
\begin{barticle}
\bauthor{\bsnm{Shi}, \binits{F.}},
\bauthor{\bsnm{Foster}, \binits{J.G.}},
\bauthor{\bsnm{Evans}, \binits{J.A.}}:
\batitle{Weaving the fabric of science: Dynamic network models of science's unfolding structure}.
\bjtitle{Social Networks}
\bvolume{43},
\bfpage{73}--\blpage{85}
(\byear{2015})
\end{barticle}
\endbibitem

\bibitem{klavans2017type}
\begin{barticle}
\bauthor{\bsnm{Klavans}, \binits{R.}},
\bauthor{\bsnm{Boyack}, \binits{K.W.}}:
\batitle{Which type of citation analysis generates the most accurate taxonomy of scientific and technical knowledge?}
\bjtitle{Journal of the Association for Information Science and Technology}
\bvolume{68}(\bissue{4}),
\bfpage{984}--\blpage{998}
(\byear{2017})
\end{barticle}
\endbibitem

\bibitem{wang2020science}
\begin{bchapter}
\bauthor{\bsnm{Wang}, \binits{D.}},
\bauthor{\bsnm{Liu}, \binits{L.}}:
\bctitle{The science of science}.
In: \bbtitle{Proceedings of the ACM/IEEE Joint Conference on Digital Libraries in 2020},
pp. \bfpage{563}--\blpage{564}
(\byear{2020})
\end{bchapter}
\endbibitem

\bibitem{wapman2022quantifying}
\begin{barticle}
\bauthor{\bsnm{Wapman}, \binits{K.H.}},
\bauthor{\bsnm{Zhang}, \binits{S.}},
\bauthor{\bsnm{Clauset}, \binits{A.}},
\bauthor{\bsnm{Larremore}, \binits{D.B.}}:
\batitle{Quantifying hierarchy and dynamics in us faculty hiring and retention}.
\bjtitle{Nature}
\bvolume{610}(\bissue{7930}),
\bfpage{120}--\blpage{127}
(\byear{2022})
\end{barticle}
\endbibitem

\bibitem{gao2024drugclip}
\begin{botherref}
\oauthor{\bsnm{Gao}, \binits{B.}},
\oauthor{\bsnm{Qiang}, \binits{B.}},
\oauthor{\bsnm{Tan}, \binits{H.}},
\oauthor{\bsnm{Jia}, \binits{Y.}},
\oauthor{\bsnm{Ren}, \binits{M.}},
\oauthor{\bsnm{Lu}, \binits{M.}},
\oauthor{\bsnm{Liu}, \binits{J.}},
\oauthor{\bsnm{Ma}, \binits{W.-Y.}},
\oauthor{\bsnm{Lan}, \binits{Y.}}:
Drugclip: Contrasive protein-molecule representation learning for virtual screening.
Advances in Neural Information Processing Systems
\textbf{36}
(2024)
\end{botherref}
\endbibitem

\bibitem{abramson2024accurate}
\begin{botherref}
\oauthor{\bsnm{Abramson}, \binits{J.}},
\oauthor{\bsnm{Adler}, \binits{J.}},
\oauthor{\bsnm{Dunger}, \binits{J.}},
\oauthor{\bsnm{Evans}, \binits{R.}},
\oauthor{\bsnm{Green}, \binits{T.}},
\oauthor{\bsnm{Pritzel}, \binits{A.}},
\oauthor{\bsnm{Ronneberger}, \binits{O.}},
\oauthor{\bsnm{Willmore}, \binits{L.}},
\oauthor{\bsnm{Ballard}, \binits{A.J.}},
\oauthor{\bsnm{Bambrick}, \binits{J.}}, et al.:
Accurate structure prediction of biomolecular interactions with alphafold 3.
Nature,
1--3
(2024)
\end{botherref}
\endbibitem

\bibitem{chang2024bidirectional}
\begin{barticle}
\bauthor{\bsnm{Chang}, \binits{J.}},
\bauthor{\bsnm{Ye}, \binits{J.C.}}:
\batitle{Bidirectional generation of structure and properties through a single molecular foundation model}.
\bjtitle{Nature Communications}
\bvolume{15}(\bissue{1}),
\bfpage{2323}
(\byear{2024})
\end{barticle}
\endbibitem

\bibitem{rzhetsky2015choosing}
\begin{barticle}
\bauthor{\bsnm{Rzhetsky}, \binits{A.}},
\bauthor{\bsnm{Foster}, \binits{J.G.}},
\bauthor{\bsnm{Foster}, \binits{I.T.}},
\bauthor{\bsnm{Evans}, \binits{J.A.}}:
\batitle{Choosing experiments to accelerate collective discovery}.
\bjtitle{Proceedings of the National Academy of Sciences}
\bvolume{112}(\bissue{47}),
\bfpage{14569}--\blpage{14574}
(\byear{2015})
\end{barticle}
\endbibitem

\bibitem{alshebli2018preeminence}
\begin{barticle}
\bauthor{\bsnm{AlShebli}, \binits{B.K.}},
\bauthor{\bsnm{Rahwan}, \binits{T.}},
\bauthor{\bsnm{Woon}, \binits{W.L.}}:
\batitle{The preeminence of ethnic diversity in scientific collaboration}.
\bjtitle{Nature communications}
\bvolume{9}(\bissue{1}),
\bfpage{5163}
(\byear{2018})
\end{barticle}
\endbibitem

\bibitem{liu2018hot}
\begin{barticle}
\bauthor{\bsnm{Liu}, \binits{L.}},
\bauthor{\bsnm{Wang}, \binits{Y.}},
\bauthor{\bsnm{Sinatra}, \binits{R.}},
\bauthor{\bsnm{Giles}, \binits{C.L.}},
\bauthor{\bsnm{Song}, \binits{C.}},
\bauthor{\bsnm{Wang}, \binits{D.}}:
\batitle{Hot streaks in artistic, cultural, and scientific careers}.
\bjtitle{Nature}
\bvolume{559}(\bissue{7714}),
\bfpage{396}--\blpage{399}
(\byear{2018})
\end{barticle}
\endbibitem

\bibitem{yin2019quantifying}
\begin{barticle}
\bauthor{\bsnm{Yin}, \binits{Y.}},
\bauthor{\bsnm{Wang}, \binits{Y.}},
\bauthor{\bsnm{Evans}, \binits{J.A.}},
\bauthor{\bsnm{Wang}, \binits{D.}}:
\batitle{Quantifying the dynamics of failure across science, startups and security}.
\bjtitle{Nature}
\bvolume{575}(\bissue{7781}),
\bfpage{190}--\blpage{194}
(\byear{2019})
\end{barticle}
\endbibitem

\bibitem{shi2023surprising}
\begin{barticle}
\bauthor{\bsnm{Shi}, \binits{F.}},
\bauthor{\bsnm{Evans}, \binits{J.}}:
\batitle{Surprising combinations of research contents and contexts are related to impact and emerge with scientific outsiders from distant disciplines}.
\bjtitle{Nature Communications}
\bvolume{14},
\bfpage{1641}
(\byear{2023})
\end{barticle}
\endbibitem

\bibitem{weston2019named}
\begin{barticle}
\bauthor{\bsnm{Weston}, \binits{L.}},
\bauthor{\bsnm{Tshitoyan}, \binits{V.}},
\bauthor{\bsnm{Dagdelen}, \binits{J.}},
\bauthor{\bsnm{Kononova}, \binits{O.}},
\bauthor{\bsnm{Trewartha}, \binits{A.}},
\bauthor{\bsnm{Persson}, \binits{K.A.}},
\bauthor{\bsnm{Ceder}, \binits{G.}},
\bauthor{\bsnm{Jain}, \binits{A.}}:
\batitle{Named entity recognition and normalization applied to large-scale information extraction from the materials science literature}.
\bjtitle{Journal of chemical information and modeling}
\bvolume{59}(\bissue{9}),
\bfpage{3692}--\blpage{3702}
(\byear{2019})
\end{barticle}
\endbibitem

\bibitem{clauset2015systematic}
\begin{barticle}
\bauthor{\bsnm{Clauset}, \binits{A.}},
\bauthor{\bsnm{Arbesman}, \binits{S.}},
\bauthor{\bsnm{Larremore}, \binits{D.B.}}:
\batitle{Systematic inequality and hierarchy in faculty hiring networks}.
\bjtitle{Science advances}
\bvolume{1}(\bissue{1}),
\bfpage{1400005}
(\byear{2015})
\end{barticle}
\endbibitem

\bibitem{xiao2016modeling}
\begin{bchapter}
\bauthor{\bsnm{Xiao}, \binits{S.}},
\bauthor{\bsnm{Yan}, \binits{J.}},
\bauthor{\bsnm{Li}, \binits{C.}},
\bauthor{\bsnm{Jin}, \binits{B.}},
\bauthor{\bsnm{Wang}, \binits{X.}},
\bauthor{\bsnm{Yang}, \binits{X.}},
\bauthor{\bsnm{Chu}, \binits{S.M.}},
\bauthor{\bsnm{Zha}, \binits{H.}}:
\bctitle{On modeling and predicting individual paper citation count over time.}
In: \bbtitle{Ijcai},
pp. \bfpage{2676}--\blpage{2682}
(\byear{2016})
\end{bchapter}
\endbibitem

\bibitem{krenn2020predicting}
\begin{barticle}
\bauthor{\bsnm{Krenn}, \binits{M.}},
\bauthor{\bsnm{Zeilinger}, \binits{A.}}:
\batitle{Predicting research trends with semantic and neural networks with an application in quantum physics}.
\bjtitle{Proceedings of the National Academy of Sciences}
\bvolume{117}(\bissue{4}),
\bfpage{1910}--\blpage{1916}
(\byear{2020})
\end{barticle}
\endbibitem

\bibitem{van2017citation}
\begin{barticle}
\bauthor{\bsnm{Van~Eck}, \binits{N.J.}},
\bauthor{\bsnm{Waltman}, \binits{L.}}:
\batitle{Citation-based clustering of publications using citnetexplorer and vosviewer}.
\bjtitle{Scientometrics}
\bvolume{111},
\bfpage{1053}--\blpage{1070}
(\byear{2017})
\end{barticle}
\endbibitem

\bibitem{hofstra2020diversity}
\begin{barticle}
\bauthor{\bsnm{Hofstra}, \binits{B.}},
\bauthor{\bsnm{Kulkarni}, \binits{V.V.}},
\bauthor{\bsnm{Munoz-Najar~Galvez}, \binits{S.}},
\bauthor{\bsnm{He}, \binits{B.}},
\bauthor{\bsnm{Jurafsky}, \binits{D.}},
\bauthor{\bsnm{McFarland}, \binits{D.A.}}:
\batitle{The diversity--innovation paradox in science}.
\bjtitle{Proceedings of the National Academy of Sciences}
\bvolume{117}(\bissue{17}),
\bfpage{9284}--\blpage{9291}
(\byear{2020})
\end{barticle}
\endbibitem

\bibitem{ghafarollahi2024sciagents}
\begin{botherref}
\oauthor{\bsnm{Ghafarollahi}, \binits{A.}},
\oauthor{\bsnm{Buehler}, \binits{M.J.}}:
Sciagents: Automating scientific discovery through multi-agent intelligent graph reasoning.
arXiv preprint arXiv:2409.05556
(2024)
\end{botherref}
\endbibitem

\bibitem{su2024two}
\begin{botherref}
\oauthor{\bsnm{Su}, \binits{H.}},
\oauthor{\bsnm{Chen}, \binits{R.}},
\oauthor{\bsnm{Tang}, \binits{S.}},
\oauthor{\bsnm{Zheng}, \binits{X.}},
\oauthor{\bsnm{Li}, \binits{J.}},
\oauthor{\bsnm{Yin}, \binits{Z.}},
\oauthor{\bsnm{Ouyang}, \binits{W.}},
\oauthor{\bsnm{Dong}, \binits{N.}}:
Two heads are better than one: A multi-agent system has the potential to improve scientific idea generation.
arXiv preprint arXiv:2410.09403
(2024)
\end{botherref}
\endbibitem

\bibitem{yang2024oasis}
\begin{botherref}
\oauthor{\bsnm{Yang}, \binits{Z.}},
\oauthor{\bsnm{Zhang}, \binits{Z.}},
\oauthor{\bsnm{Zheng}, \binits{Z.}},
\oauthor{\bsnm{Jiang}, \binits{Y.}},
\oauthor{\bsnm{Gan}, \binits{Z.}},
\oauthor{\bsnm{Wang}, \binits{Z.}},
\oauthor{\bsnm{Ling}, \binits{Z.}},
\oauthor{\bsnm{Chen}, \binits{J.}},
\oauthor{\bsnm{Ma}, \binits{M.}},
\oauthor{\bsnm{Dong}, \binits{B.}}, et al.:
Oasis: Open agents social interaction simulations on one million agents.
arXiv preprint arXiv:2411.11581
(2024)
\end{botherref}
\endbibitem

\bibitem{li2024ai}
\begin{botherref}
\oauthor{\bsnm{Li}, \binits{Z.}},
\oauthor{\bsnm{Song}, \binits{P.}},
\oauthor{\bsnm{Li}, \binits{G.}},
\oauthor{\bsnm{Han}, \binits{Y.}},
\oauthor{\bsnm{Ren}, \binits{X.}},
\oauthor{\bsnm{Bai}, \binits{L.}},
\oauthor{\bsnm{Su}, \binits{J.}}:
Ai energized hydrogel design, optimization and application in biomedicine.
Materials Today Bio,
101014
(2024)
\end{botherref}
\endbibitem

\bibitem{ofosu2024artificial}
\begin{botherref}
\oauthor{\bsnm{Ofosu-Ampong}, \binits{K.}}:
Artificial intelligence research: A review on dominant themes, methods, frameworks and future research directions.
Telematics and Informatics Reports,
100127
(2024)
\end{botherref}
\endbibitem

\bibitem{iacopini2018network}
\begin{barticle}
\bauthor{\bsnm{Iacopini}, \binits{I.}},
\bauthor{\bsnm{Milojevi{\'c}}, \binits{S.}},
\bauthor{\bsnm{Latora}, \binits{V.}}:
\batitle{Network dynamics of innovation processes}.
\bjtitle{Physical review letters}
\bvolume{120}(\bissue{4}),
\bfpage{048301}
(\byear{2018})
\end{barticle}
\endbibitem

\bibitem{bolt2021educating}
\begin{barticle}
\bauthor{\bsnm{Bolt}, \binits{T.}},
\bauthor{\bsnm{Nomi}, \binits{J.S.}},
\bauthor{\bsnm{Bzdok}, \binits{D.}},
\bauthor{\bsnm{Uddin}, \binits{L.Q.}}:
\batitle{Educating the future generation of researchers: A cross-disciplinary survey of trends in analysis methods}.
\bjtitle{PLoS biology}
\bvolume{19}(\bissue{7}),
\bfpage{3001313}
(\byear{2021})
\end{barticle}
\endbibitem

\bibitem{borner2018forecasting}
\begin{barticle}
\bauthor{\bsnm{B{\"o}rner}, \binits{K.}},
\bauthor{\bsnm{Rouse}, \binits{W.B.}},
\bauthor{\bsnm{Trunfio}, \binits{P.}},
\bauthor{\bsnm{Stanley}, \binits{H.E.}}:
\batitle{Forecasting innovations in science, technology, and education}.
\bjtitle{Proceedings of the National Academy of Sciences}
\bvolume{115}(\bissue{50}),
\bfpage{12573}--\blpage{12581}
(\byear{2018})
\end{barticle}
\endbibitem

\bibitem{zhou2020graph}
\begin{barticle}
\bauthor{\bsnm{Zhou}, \binits{J.}},
\bauthor{\bsnm{Cui}, \binits{G.}},
\bauthor{\bsnm{Hu}, \binits{S.}},
\bauthor{\bsnm{Zhang}, \binits{Z.}},
\bauthor{\bsnm{Yang}, \binits{C.}},
\bauthor{\bsnm{Liu}, \binits{Z.}},
\bauthor{\bsnm{Wang}, \binits{L.}},
\bauthor{\bsnm{Li}, \binits{C.}},
\bauthor{\bsnm{Sun}, \binits{M.}}:
\batitle{Graph neural networks: A review of methods and applications}.
\bjtitle{AI open}
\bvolume{1},
\bfpage{57}--\blpage{81}
(\byear{2020})
\end{barticle}
\endbibitem

\bibitem{wang2019early}
\begin{barticle}
\bauthor{\bsnm{Wang}, \binits{Y.}},
\bauthor{\bsnm{Jones}, \binits{B.F.}},
\bauthor{\bsnm{Wang}, \binits{D.}}:
\batitle{Early-career setback and future career impact}.
\bjtitle{Nature communications}
\bvolume{10}(\bissue{1}),
\bfpage{4331}
(\byear{2019})
\end{barticle}
\endbibitem

\bibitem{wu2019large}
\begin{barticle}
\bauthor{\bsnm{Wu}, \binits{L.}},
\bauthor{\bsnm{Wang}, \binits{D.}},
\bauthor{\bsnm{Evans}, \binits{J.A.}}:
\batitle{Large teams develop and small teams disrupt science and technology}.
\bjtitle{Nature}
\bvolume{566},
\bfpage{378}--\blpage{382}
(\byear{2019})
\end{barticle}
\endbibitem

\bibitem{wuchty2007increasing}
\begin{barticle}
\bauthor{\bsnm{Wuchty}, \binits{S.}},
\bauthor{\bsnm{Jones}, \binits{B.F.}},
\bauthor{\bsnm{Uzzi}, \binits{B.}}:
\batitle{The increasing dominance of teams in production of knowledge}.
\bjtitle{Science}
\bvolume{316}(\bissue{5827}),
\bfpage{1036}--\blpage{1039}
(\byear{2007})
\end{barticle}
\endbibitem

\bibitem{yang2022gender}
\begin{barticle}
\bauthor{\bsnm{Yang}, \binits{Y.}},
\bauthor{\bsnm{Tian}, \binits{T.Y.}},
\bauthor{\bsnm{Woodruff}, \binits{T.K.}},
\bauthor{\bsnm{Jones}, \binits{B.F.}},
\bauthor{\bsnm{Uzzi}, \binits{B.}}:
\batitle{Gender-diverse teams produce more novel and higher-impact scientific ideas}.
\bjtitle{Proceedings of the National Academy of Sciences}
\bvolume{119}(\bissue{36}),
\bfpage{2200841119}
(\byear{2022})
\end{barticle}
\endbibitem

\bibitem{liu2023non}
\begin{barticle}
\bauthor{\bsnm{Liu}, \binits{F.}},
\bauthor{\bsnm{Rahwan}, \binits{T.}},
\bauthor{\bsnm{AlShebli}, \binits{B.}}:
\batitle{Non-white scientists appear on fewer editorial boards, spend more time under review, and receive fewer citations}.
\bjtitle{Proceedings of the National Academy of Sciences}
\bvolume{120}(\bissue{13}),
\bfpage{2215324120}
(\byear{2023})
\end{barticle}
\endbibitem

\bibitem{gomez2022leading}
\begin{barticle}
\bauthor{\bsnm{Gomez}, \binits{C.J.}},
\bauthor{\bsnm{Herman}, \binits{A.C.}},
\bauthor{\bsnm{Parigi}, \binits{P.}}:
\batitle{Leading countries in global science increasingly receive more citations than other countries doing similar research}.
\bjtitle{Nature Human Behaviour}
\bvolume{6}(\bissue{7}),
\bfpage{919}--\blpage{929}
(\byear{2022})
\end{barticle}
\endbibitem

\bibitem{guimera2005team}
\begin{barticle}
\bauthor{\bsnm{Guimera}, \binits{R.}},
\bauthor{\bsnm{Uzzi}, \binits{B.}},
\bauthor{\bsnm{Spiro}, \binits{J.}},
\bauthor{\bsnm{Amaral}, \binits{L.A.N.}}:
\batitle{Team assembly mechanisms determine collaboration network structure and team performance}.
\bjtitle{Science}
\bvolume{308}(\bissue{5722}),
\bfpage{697}--\blpage{702}
(\byear{2005})
\end{barticle}
\endbibitem

\bibitem{fernandez2017insight}
\begin{barticle}
\bauthor{\bsnm{Fern{\'a}ndez}, \binits{A.}},
\bauthor{\bparticle{del} \bsnm{R{\'\i}o}, \binits{S.}},
\bauthor{\bsnm{Chawla}, \binits{N.V.}},
\bauthor{\bsnm{Herrera}, \binits{F.}}:
\batitle{An insight into imbalanced big data classification: outcomes and challenges}.
\bjtitle{Complex \& Intelligent Systems}
\bvolume{3},
\bfpage{105}--\blpage{120}
(\byear{2017})
\end{barticle}
\endbibitem

\bibitem{kaur2019systematic}
\begin{barticle}
\bauthor{\bsnm{Kaur}, \binits{H.}},
\bauthor{\bsnm{Pannu}, \binits{H.S.}},
\bauthor{\bsnm{Malhi}, \binits{A.K.}}:
\batitle{A systematic review on imbalanced data challenges in machine learning: Applications and solutions}.
\bjtitle{ACM computing surveys (CSUR)}
\bvolume{52}(\bissue{4}),
\bfpage{1}--\blpage{36}
(\byear{2019})
\end{barticle}
\endbibitem

\bibitem{leevy2018survey}
\begin{barticle}
\bauthor{\bsnm{Leevy}, \binits{J.L.}},
\bauthor{\bsnm{Khoshgoftaar}, \binits{T.M.}},
\bauthor{\bsnm{Bauder}, \binits{R.A.}},
\bauthor{\bsnm{Seliya}, \binits{N.}}:
\batitle{A survey on addressing high-class imbalance in big data}.
\bjtitle{Journal of Big Data}
\bvolume{5}(\bissue{1}),
\bfpage{1}--\blpage{30}
(\byear{2018})
\end{barticle}
\endbibitem

\bibitem{johnson2019survey}
\begin{barticle}
\bauthor{\bsnm{Johnson}, \binits{J.M.}},
\bauthor{\bsnm{Khoshgoftaar}, \binits{T.M.}}:
\batitle{Survey on deep learning with class imbalance}.
\bjtitle{Journal of big data}
\bvolume{6}(\bissue{1}),
\bfpage{1}--\blpage{54}
(\byear{2019})
\end{barticle}
\endbibitem

\bibitem{liu2023data}
\begin{barticle}
\bauthor{\bsnm{Liu}, \binits{L.}},
\bauthor{\bsnm{Jones}, \binits{B.F.}},
\bauthor{\bsnm{Uzzi}, \binits{B.}},
\bauthor{\bsnm{Wang}, \binits{D.}}:
\batitle{Data, measurement and empirical methods in the science of science}.
\bjtitle{Nature human behaviour}
\bvolume{7}(\bissue{7}),
\bfpage{1046}--\blpage{1058}
(\byear{2023})
\end{barticle}
\endbibitem

\bibitem{achiam2023gpt}
\begin{botherref}
\oauthor{\bsnm{Achiam}, \binits{J.}},
\oauthor{\bsnm{Adler}, \binits{S.}},
\oauthor{\bsnm{Agarwal}, \binits{S.}},
\oauthor{\bsnm{Ahmad}, \binits{L.}},
\oauthor{\bsnm{Akkaya}, \binits{I.}},
\oauthor{\bsnm{Aleman}, \binits{F.L.}},
\oauthor{\bsnm{Almeida}, \binits{D.}},
\oauthor{\bsnm{Altenschmidt}, \binits{J.}},
\oauthor{\bsnm{Altman}, \binits{S.}},
\oauthor{\bsnm{Anadkat}, \binits{S.}}, et al.:
Gpt-4 technical report.
arXiv preprint arXiv:2303.08774
(2023)
\end{botherref}
\endbibitem

\bibitem{dubey2024llama}
\begin{botherref}
\oauthor{\bsnm{Dubey}, \binits{A.}},
\oauthor{\bsnm{Jauhri}, \binits{A.}},
\oauthor{\bsnm{Pandey}, \binits{A.}},
\oauthor{\bsnm{Kadian}, \binits{A.}},
\oauthor{\bsnm{Al-Dahle}, \binits{A.}},
\oauthor{\bsnm{Letman}, \binits{A.}},
\oauthor{\bsnm{Mathur}, \binits{A.}},
\oauthor{\bsnm{Schelten}, \binits{A.}},
\oauthor{\bsnm{Yang}, \binits{A.}},
\oauthor{\bsnm{Fan}, \binits{A.}}, et al.:
The llama 3 herd of models.
arXiv preprint arXiv:2407.21783
(2024)
\end{botherref}
\endbibitem

\bibitem{yang2024qwen2}
\begin{botherref}
\oauthor{\bsnm{Yang}, \binits{A.}},
\oauthor{\bsnm{Yang}, \binits{B.}},
\oauthor{\bsnm{Zhang}, \binits{B.}},
\oauthor{\bsnm{Hui}, \binits{B.}},
\oauthor{\bsnm{Zheng}, \binits{B.}},
\oauthor{\bsnm{Yu}, \binits{B.}},
\oauthor{\bsnm{Li}, \binits{C.}},
\oauthor{\bsnm{Liu}, \binits{D.}},
\oauthor{\bsnm{Huang}, \binits{F.}},
\oauthor{\bsnm{Wei}, \binits{H.}}, et al.:
Qwen2. 5 technical report.
arXiv preprint arXiv:2412.15115
(2024)
\end{botherref}
\endbibitem

\bibitem{mehrabi2021survey}
\begin{barticle}
\bauthor{\bsnm{Mehrabi}, \binits{N.}},
\bauthor{\bsnm{Morstatter}, \binits{F.}},
\bauthor{\bsnm{Saxena}, \binits{N.}},
\bauthor{\bsnm{Lerman}, \binits{K.}},
\bauthor{\bsnm{Galstyan}, \binits{A.}}:
\batitle{A survey on bias and fairness in machine learning}.
\bjtitle{ACM computing surveys (CSUR)}
\bvolume{54}(\bissue{6}),
\bfpage{1}--\blpage{35}
(\byear{2021})
\end{barticle}
\endbibitem

\bibitem{sinha2015overview}
\begin{bchapter}
\bauthor{\bsnm{Sinha}, \binits{A.}},
\bauthor{\bsnm{Shen}, \binits{Z.}},
\bauthor{\bsnm{Song}, \binits{Y.}},
\bauthor{\bsnm{Ma}, \binits{H.}},
\bauthor{\bsnm{Eide}, \binits{D.}},
\bauthor{\bsnm{Hsu}, \binits{B.-J.}},
\bauthor{\bsnm{Wang}, \binits{K.}}:
\bctitle{An overview of microsoft academic service (mas) and applications}.
In: \bbtitle{Proceedings of the 24th International Conference on World Wide Web},
pp. \bfpage{243}--\blpage{246}
(\byear{2015})
\end{bchapter}
\endbibitem

\bibitem{zhang2022oag}
\begin{barticle}
\bauthor{\bsnm{Zhang}, \binits{F.}},
\bauthor{\bsnm{Liu}, \binits{X.}},
\bauthor{\bsnm{Tang}, \binits{J.}},
\bauthor{\bsnm{Dong}, \binits{Y.}},
\bauthor{\bsnm{Yao}, \binits{P.}},
\bauthor{\bsnm{Zhang}, \binits{J.}},
\bauthor{\bsnm{Gu}, \binits{X.}},
\bauthor{\bsnm{Wang}, \binits{Y.}},
\bauthor{\bsnm{Kharlamov}, \binits{E.}},
\bauthor{\bsnm{Shao}, \binits{B.}}, \betal:
\batitle{Oag: Linking entities across large-scale heterogeneous knowledge graphs}.
\bjtitle{IEEE Transactions on Knowledge and Data Engineering}
\bvolume{35}(\bissue{9}),
\bfpage{9225}--\blpage{9239}
(\byear{2022})
\end{barticle}
\endbibitem

\bibitem{lin2023sciscinet}
\begin{barticle}
\bauthor{\bsnm{Lin}, \binits{Z.}},
\bauthor{\bsnm{Yin}, \binits{Y.}},
\bauthor{\bsnm{Liu}, \binits{L.}},
\bauthor{\bsnm{Wang}, \binits{D.}}:
\batitle{Sciscinet: A large-scale open data lake for the science of science research}.
\bjtitle{Scientific Data}
\bvolume{10}(\bissue{1}),
\bfpage{315}
(\byear{2023})
\end{barticle}
\endbibitem

\bibitem{scatiggio2020tackling}
\begin{botherref}
\oauthor{\bsnm{Scatiggio}, \binits{V.}}:
Tackling the issue of bias in artificial intelligence to design ai-driven fair and inclusive service systems. how human biases are breaching into ai algorithms, with severe impacts on individuals and societies, and what designers can do to face this phenomenon and change for the better
(2020)
\end{botherref}
\endbibitem

\bibitem{prabhakaran2021releasing}
\begin{botherref}
\oauthor{\bsnm{Prabhakaran}, \binits{V.}},
\oauthor{\bsnm{Davani}, \binits{A.M.}},
\oauthor{\bsnm{Diaz}, \binits{M.}}:
On releasing annotator-level labels and information in datasets.
arXiv preprint arXiv:2110.05699
(2021)
\end{botherref}
\endbibitem

\bibitem{norling2017informal}
\begin{botherref}
\oauthor{\bsnm{Norling}, \binits{E.}},
\oauthor{\bsnm{Edmonds}, \binits{B.}},
\oauthor{\bsnm{Meyer}, \binits{R.}}:
Informal approaches to developing simulation models.
Simulating Social Complexity: A Handbook,
61--79
(2017)
\end{botherref}
\endbibitem

\bibitem{gao2024large}
\begin{barticle}
\bauthor{\bsnm{Gao}, \binits{C.}},
\bauthor{\bsnm{Lan}, \binits{X.}},
\bauthor{\bsnm{Li}, \binits{N.}},
\bauthor{\bsnm{Yuan}, \binits{Y.}},
\bauthor{\bsnm{Ding}, \binits{J.}},
\bauthor{\bsnm{Zhou}, \binits{Z.}},
\bauthor{\bsnm{Xu}, \binits{F.}},
\bauthor{\bsnm{Li}, \binits{Y.}}:
\batitle{Large language models empowered agent-based modeling and simulation: A survey and perspectives}.
\bjtitle{Humanities and Social Sciences Communications}
\bvolume{11}(\bissue{1}),
\bfpage{1}--\blpage{24}
(\byear{2024})
\end{barticle}
\endbibitem

\bibitem{khaleghian2023calibrating}
\begin{bchapter}
\bauthor{\bsnm{Khaleghian}, \binits{S.}},
\bauthor{\bsnm{Neema}, \binits{H.}},
\bauthor{\bsnm{Sartipi}, \binits{M.}},
\bauthor{\bsnm{Tran}, \binits{T.}},
\bauthor{\bsnm{Sen}, \binits{R.}},
\bauthor{\bsnm{Dubey}, \binits{A.}}:
\bctitle{Calibrating real-world city traffic simulation model using vehicle speed data}.
In: \bbtitle{2023 IEEE International Conference on Smart Computing (SMARTCOMP)},
pp. \bfpage{303}--\blpage{308}
(\byear{2023}).
\bcomment{IEEE}
\end{bchapter}
\endbibitem

\bibitem{schulze2017agent}
\begin{botherref}
\oauthor{\bsnm{Schulze}, \binits{J.}},
\oauthor{\bsnm{M{\"u}ller}, \binits{B.}},
\oauthor{\bsnm{Groeneveld}, \binits{J.}},
\oauthor{\bsnm{Grimm}, \binits{V.}}:
Agent-based modelling of social-ecological systems: achievements, challenges, and a way forward.
Journal of Artificial Societies and Social Simulation
\textbf{20}(2)
(2017)
\end{botherref}
\endbibitem

\bibitem{an2021challenges}
\begin{barticle}
\bauthor{\bsnm{An}, \binits{L.}},
\bauthor{\bsnm{Grimm}, \binits{V.}},
\bauthor{\bsnm{Sullivan}, \binits{A.}},
\bauthor{\bsnm{Turner~Ii}, \binits{B.}},
\bauthor{\bsnm{Malleson}, \binits{N.}},
\bauthor{\bsnm{Heppenstall}, \binits{A.}},
\bauthor{\bsnm{Vincenot}, \binits{C.}},
\bauthor{\bsnm{Robinson}, \binits{D.}},
\bauthor{\bsnm{Ye}, \binits{X.}},
\bauthor{\bsnm{Liu}, \binits{J.}}, \betal:
\batitle{Challenges, tasks, and opportunities in modeling agent-based complex systems}.
\bjtitle{Ecological Modelling}
\bvolume{457},
\bfpage{109685}
(\byear{2021})
\end{barticle}
\endbibitem

\bibitem{ebadi2015receive}
\begin{barticle}
\bauthor{\bsnm{Ebadi}, \binits{A.}},
\bauthor{\bsnm{Schiffauerova}, \binits{A.}}:
\batitle{How to receive more funding for your research? get connected to the right people!}
\bjtitle{PloS one}
\bvolume{10}(\bissue{7}),
\bfpage{0133061}
(\byear{2015})
\end{barticle}
\endbibitem

\bibitem{ronda2018evolutions}
\begin{barticle}
\bauthor{\bsnm{Ronda-Pupo}, \binits{G.A.}},
\bauthor{\bsnm{Pham}, \binits{T.}}:
\batitle{The evolutions of the rich get richer and the fit get richer phenomena in scholarly networks: The case of the strategic management journal}.
\bjtitle{Scientometrics}
\bvolume{116}(\bissue{1}),
\bfpage{363}--\blpage{383}
(\byear{2018})
\end{barticle}
\endbibitem

\bibitem{katz2020metrics}
\begin{barticle}
\bauthor{\bsnm{Katz}, \binits{Y.}},
\bauthor{\bsnm{Matter}, \binits{U.}}:
\batitle{Metrics of inequality: The concentration of resources in the us biomedical elite}.
\bjtitle{Science as Culture}
\bvolume{29}(\bissue{4}),
\bfpage{475}--\blpage{502}
(\byear{2020})
\end{barticle}
\endbibitem

\bibitem{qian2024chat}
\begin{bchapter}
\bauthor{\bsnm{Qian}, \binits{C.}},
\bauthor{\bsnm{Liu}, \binits{W.}},
\bauthor{\bsnm{Liu}, \binits{H.}},
\bauthor{\bsnm{Chen}, \binits{N.}},
\bauthor{\bsnm{Dang}, \binits{Y.}},
\bauthor{\bsnm{Li}, \binits{J.}},
\bauthor{\bsnm{Yang}, \binits{C.}},
\bauthor{\bsnm{Chen}, \binits{W.}},
\bauthor{\bsnm{Su}, \binits{Y.}},
\bauthor{\bsnm{Cong}, \binits{X.}},
\bauthor{\bsnm{Xu}, \binits{J.}},
\bauthor{\bsnm{Li}, \binits{D.}},
\bauthor{\bsnm{Liu}, \binits{Z.}},
\bauthor{\bsnm{Sun}, \binits{M.}}:
\bctitle{Chatdev: Communicative agents for software development}.
In: \bbtitle{Proceedings of the 62nd Annual Meeting of the Association for Computational Linguistics},
pp. \bfpage{15174}--\blpage{15186}
(\byear{2024})
\end{bchapter}
\endbibitem

\bibitem{guo2024large}
\begin{botherref}
\oauthor{\bsnm{Guo}, \binits{T.}},
\oauthor{\bsnm{Chen}, \binits{X.}},
\oauthor{\bsnm{Wang}, \binits{Y.}},
\oauthor{\bsnm{Chang}, \binits{R.}},
\oauthor{\bsnm{Pei}, \binits{S.}},
\oauthor{\bsnm{Chawla}, \binits{N.V.}},
\oauthor{\bsnm{Wiest}, \binits{O.}},
\oauthor{\bsnm{Zhang}, \binits{X.}}:
Large language model based multi-agents: A survey of progress and challenges.
arXiv preprint arXiv:2402.01680
(2024)
\end{botherref}
\endbibitem

\bibitem{aldrich2013unsupervised}
\begin{bbook}
\bauthor{\bsnm{Aldrich}, \binits{C.}},
\bauthor{\bsnm{Auret}, \binits{L.}}:
\bbtitle{Unsupervised Process Monitoring and Fault Diagnosis with Machine Learning Methods}
vol. \bseriesno{16}.
\bpublisher{Springer}, \blocation{???}
(\byear{2013})
\end{bbook}
\endbibitem

\bibitem{chen2024philosopher}
\begin{botherref}
\oauthor{\bsnm{Chen}, \binits{Q.}},
\oauthor{\bsnm{Ho}, \binits{Y.-J.I.}},
\oauthor{\bsnm{Sun}, \binits{P.}},
\oauthor{\bsnm{Wang}, \binits{D.}}:
The philosopher’s stone for science--the catalyst change of ai for scientific creativity.
Pin and Wang, Dashun, The Philosopher’s Stone for Science--The Catalyst Change of AI for Scientific Creativity (March 5, 2024)
(2024)
\end{botherref}
\endbibitem

\bibitem{balasubramaniam2024road}
\begin{botherref}
\oauthor{\bsnm{Balasubramaniam}, \binits{S.}},
\oauthor{\bsnm{Chirchi}, \binits{V.}},
\oauthor{\bsnm{Kadry}, \binits{S.}},
\oauthor{\bsnm{Agoramoorthy}, \binits{M.}},
\oauthor{\bsnm{Gururama}, \binits{S.P.}},
\oauthor{\bsnm{Satheesh}, \binits{K.K.}},
\oauthor{\bsnm{Sivakumar}, \binits{T.}}:
The road ahead: Emerging trends, unresolved issues, and concluding remarks in generative ai—a comprehensive review.
International Journal of Intelligent Systems
\textbf{2024}
(2024)
\end{botherref}
\endbibitem

\bibitem{lissack2024navigating}
\begin{botherref}
\oauthor{\bsnm{Lissack}, \binits{M.}},
\oauthor{\bsnm{Meagher}, \binits{B.}}:
Navigating the future of large language models in scientific research: Opportunities, challenges, and ethical considerations.
Challenges, and Ethical Considerations (September 02, 2024)
(2024)
\end{botherref}
\endbibitem

\bibitem{lu2024ai}
\begin{botherref}
\oauthor{\bsnm{Lu}, \binits{C.}},
\oauthor{\bsnm{Lu}, \binits{C.}},
\oauthor{\bsnm{Lange}, \binits{R.T.}},
\oauthor{\bsnm{Foerster}, \binits{J.}},
\oauthor{\bsnm{Clune}, \binits{J.}},
\oauthor{\bsnm{Ha}, \binits{D.}}:
The ai scientist: Towards fully automated open-ended scientific discovery.
arXiv preprint arXiv:2408.06292
(2024)
\end{botherref}
\endbibitem

\bibitem{meadows2024localvaluebench}
\begin{botherref}
\oauthor{\bsnm{Meadows}, \binits{G.I.}},
\oauthor{\bsnm{Lau}, \binits{N.W.L.}},
\oauthor{\bsnm{Susanto}, \binits{E.A.}},
\oauthor{\bsnm{Yu}, \binits{C.L.}},
\oauthor{\bsnm{Paul}, \binits{A.}}:
Localvaluebench: A collaboratively built and extensible benchmark for evaluating localized value alignment and ethical safety in large language models.
arXiv preprint arXiv:2408.01460
(2024)
\end{botherref}
\endbibitem

\bibitem{ji2024moralbench}
\begin{botherref}
\oauthor{\bsnm{Ji}, \binits{J.}},
\oauthor{\bsnm{Chen}, \binits{Y.}},
\oauthor{\bsnm{Jin}, \binits{M.}},
\oauthor{\bsnm{Xu}, \binits{W.}},
\oauthor{\bsnm{Hua}, \binits{W.}},
\oauthor{\bsnm{Zhang}, \binits{Y.}}:
Moralbench: Moral evaluation of llms.
arXiv preprint arXiv:2406.04428
(2024)
\end{botherref}
\endbibitem

\bibitem{hassija2024interpreting}
\begin{barticle}
\bauthor{\bsnm{Hassija}, \binits{V.}},
\bauthor{\bsnm{Chamola}, \binits{V.}},
\bauthor{\bsnm{Mahapatra}, \binits{A.}},
\bauthor{\bsnm{Singal}, \binits{A.}},
\bauthor{\bsnm{Goel}, \binits{D.}},
\bauthor{\bsnm{Huang}, \binits{K.}},
\bauthor{\bsnm{Scardapane}, \binits{S.}},
\bauthor{\bsnm{Spinelli}, \binits{I.}},
\bauthor{\bsnm{Mahmud}, \binits{M.}},
\bauthor{\bsnm{Hussain}, \binits{A.}}:
\batitle{Interpreting black-box models: a review on explainable artificial intelligence}.
\bjtitle{Cognitive Computation}
\bvolume{16}(\bissue{1}),
\bfpage{45}--\blpage{74}
(\byear{2024})
\end{barticle}
\endbibitem

\bibitem{reddy2025towards}
\begin{bchapter}
\bauthor{\bsnm{Reddy}, \binits{C.K.}},
\bauthor{\bsnm{Shojaee}, \binits{P.}}:
\bctitle{Towards scientific discovery with generative ai: Progress, opportunities, and challenges}.
In: \bbtitle{Proceedings of the AAAI Conference on Artificial Intelligence},
vol. \bseriesno{39},
pp. \bfpage{28601}--\blpage{28609}
(\byear{2025})
\end{bchapter}
\endbibitem

\bibitem{sonnenwald2007scientific}
\begin{barticle}
\bauthor{\bsnm{Sonnenwald}, \binits{D.H.}}:
\batitle{Scientific collaboration}.
\bjtitle{Annu. Rev. Inf. Sci. Technol.}
\bvolume{41}(\bissue{1}),
\bfpage{643}--\blpage{681}
(\byear{2007})
\end{barticle}
\endbibitem

\bibitem{petersen2014causal}
\begin{barticle}
\bauthor{\bsnm{Petersen}, \binits{M.L.}},
\bauthor{\bparticle{van~der} \bsnm{Laan}, \binits{M.J.}}:
\batitle{Causal models and learning from data: integrating causal modeling and statistical estimation}.
\bjtitle{Epidemiology}
\bvolume{25}(\bissue{3}),
\bfpage{418}--\blpage{426}
(\byear{2014})
\end{barticle}
\endbibitem

\bibitem{feuerriegel2024causal}
\begin{barticle}
\bauthor{\bsnm{Feuerriegel}, \binits{S.}},
\bauthor{\bsnm{Frauen}, \binits{D.}},
\bauthor{\bsnm{Melnychuk}, \binits{V.}},
\bauthor{\bsnm{Schweisthal}, \binits{J.}},
\bauthor{\bsnm{Hess}, \binits{K.}},
\bauthor{\bsnm{Curth}, \binits{A.}},
\bauthor{\bsnm{Bauer}, \binits{S.}},
\bauthor{\bsnm{Kilbertus}, \binits{N.}},
\bauthor{\bsnm{Kohane}, \binits{I.S.}},
\bauthor{\bparticle{van~der} \bsnm{Schaar}, \binits{M.}}:
\batitle{Causal machine learning for predicting treatment outcomes}.
\bjtitle{Nature Medicine}
\bvolume{30}(\bissue{4}),
\bfpage{958}--\blpage{968}
(\byear{2024})
\end{barticle}
\endbibitem

\bibitem{dwivedi2023explainable}
\begin{barticle}
\bauthor{\bsnm{Dwivedi}, \binits{R.}},
\bauthor{\bsnm{Dave}, \binits{D.}},
\bauthor{\bsnm{Naik}, \binits{H.}},
\bauthor{\bsnm{Singhal}, \binits{S.}},
\bauthor{\bsnm{Omer}, \binits{R.}},
\bauthor{\bsnm{Patel}, \binits{P.}},
\bauthor{\bsnm{Qian}, \binits{B.}},
\bauthor{\bsnm{Wen}, \binits{Z.}},
\bauthor{\bsnm{Shah}, \binits{T.}},
\bauthor{\bsnm{Morgan}, \binits{G.}}, \betal:
\batitle{Explainable ai (xai): Core ideas, techniques, and solutions}.
\bjtitle{ACM Computing Surveys}
\bvolume{55}(\bissue{9}),
\bfpage{1}--\blpage{33}
(\byear{2023})
\end{barticle}
\endbibitem

\bibitem{longo2024explainable}
\begin{barticle}
\bauthor{\bsnm{Longo}, \binits{L.}},
\bauthor{\bsnm{Brcic}, \binits{M.}},
\bauthor{\bsnm{Cabitza}, \binits{F.}},
\bauthor{\bsnm{Choi}, \binits{J.}},
\bauthor{\bsnm{Confalonieri}, \binits{R.}},
\bauthor{\bsnm{Del~Ser}, \binits{J.}},
\bauthor{\bsnm{Guidotti}, \binits{R.}},
\bauthor{\bsnm{Hayashi}, \binits{Y.}},
\bauthor{\bsnm{Herrera}, \binits{F.}},
\bauthor{\bsnm{Holzinger}, \binits{A.}}, \betal:
\batitle{Explainable artificial intelligence (xai) 2.0: A manifesto of open challenges and interdisciplinary research directions}.
\bjtitle{Information Fusion}
\bvolume{106},
\bfpage{102301}
(\byear{2024})
\end{barticle}
\endbibitem

\bibitem{king2011comparative}
\begin{botherref}
\oauthor{\bsnm{King}, \binits{G.}},
\oauthor{\bsnm{Nielsen}, \binits{R.}},
\oauthor{\bsnm{Coberley}, \binits{C.}},
\oauthor{\bsnm{Pope}, \binits{J.E.}},
\oauthor{\bsnm{Wells}, \binits{A.}}:
Comparative effectiveness of matching methods for causal inference.
Unpublished manuscript, Institute for Quantitative Social Science, Harvard University, Cambridge, MA
(2011)
\end{botherref}
\endbibitem

\bibitem{imbens2015causal}
\begin{bbook}
\bauthor{\bsnm{Imbens}, \binits{G.W.}},
\bauthor{\bsnm{Rubin}, \binits{D.B.}}:
\bbtitle{Causal Inference in Statistics, Social, and Biomedical Sciences}.
\bpublisher{Cambridge university press}, \blocation{???}
(\byear{2015})
\end{bbook}
\endbibitem

\bibitem{de2014applications}
\begin{barticle}
\bauthor{\bsnm{De~Carvalho}, \binits{J.}},
\bauthor{\bsnm{Chima}, \binits{F.O.}}:
\batitle{Applications of structural equation modeling in social sciences research}.
\bjtitle{American International Journal of Contemporary Research}
\bvolume{4}(\bissue{1}),
\bfpage{6}--\blpage{11}
(\byear{2014})
\end{barticle}
\endbibitem

\bibitem{khan2019methodological}
\begin{barticle}
\bauthor{\bsnm{Khan}, \binits{G.F.}},
\bauthor{\bsnm{Sarstedt}, \binits{M.}},
\bauthor{\bsnm{Shiau}, \binits{W.-L.}},
\bauthor{\bsnm{Hair}, \binits{J.F.}},
\bauthor{\bsnm{Ringle}, \binits{C.M.}},
\bauthor{\bsnm{Fritze}, \binits{M.P.}}:
\batitle{Methodological research on partial least squares structural equation modeling (pls-sem) an analysis based on social network approaches}.
\bjtitle{Internet Research}
\bvolume{29}(\bissue{3}),
\bfpage{407}--\blpage{429}
(\byear{2019})
\end{barticle}
\endbibitem

\bibitem{leist2022mapping}
\begin{barticle}
\bauthor{\bsnm{Leist}, \binits{A.K.}},
\bauthor{\bsnm{Klee}, \binits{M.}},
\bauthor{\bsnm{Kim}, \binits{J.H.}},
\bauthor{\bsnm{Rehkopf}, \binits{D.H.}},
\bauthor{\bsnm{Bordas}, \binits{S.P.}},
\bauthor{\bsnm{Muniz-Terrera}, \binits{G.}},
\bauthor{\bsnm{Wade}, \binits{S.}}:
\batitle{Mapping of machine learning approaches for description, prediction, and causal inference in the social and health sciences}.
\bjtitle{Science Advances}
\bvolume{8}(\bissue{42}),
\bfpage{1942}
(\byear{2022})
\end{barticle}
\endbibitem

\bibitem{qi2024large}
\begin{botherref}
\oauthor{\bsnm{Qi}, \binits{B.}},
\oauthor{\bsnm{Zhang}, \binits{K.}},
\oauthor{\bsnm{Tian}, \binits{K.}},
\oauthor{\bsnm{Li}, \binits{H.}},
\oauthor{\bsnm{Chen}, \binits{Z.-R.}},
\oauthor{\bsnm{Zeng}, \binits{S.}},
\oauthor{\bsnm{Hua}, \binits{E.}},
\oauthor{\bsnm{Jinfang}, \binits{H.}},
\oauthor{\bsnm{Zhou}, \binits{B.}}:
Large language models as biomedical hypothesis generators: A comprehensive evaluation.
arXiv preprint arXiv:2407.08940
(2024)
\end{botherref}
\endbibitem

\bibitem{ambekar2009name}
\begin{bchapter}
\bauthor{\bsnm{Ambekar}, \binits{A.}},
\bauthor{\bsnm{Ward}, \binits{C.}},
\bauthor{\bsnm{Mohammed}, \binits{J.}},
\bauthor{\bsnm{Male}, \binits{S.}},
\bauthor{\bsnm{Skiena}, \binits{S.}}:
\bctitle{Name-ethnicity classification from open sources}.
In: \bbtitle{Proceedings of the 15th ACM SIGKDD International Conference on Knowledge Discovery and Data Mining},
pp. \bfpage{49}--\blpage{58}
(\byear{2009})
\end{bchapter}
\endbibitem

\bibitem{li2019early}
\begin{barticle}
\bauthor{\bsnm{Li}, \binits{W.}},
\bauthor{\bsnm{Aste}, \binits{T.}},
\bauthor{\bsnm{Caccioli}, \binits{F.}},
\bauthor{\bsnm{Livan}, \binits{G.}}:
\batitle{Early coauthorship with top scientists predicts success in academic careers}.
\bjtitle{Nature communications}
\bvolume{10}(\bissue{1}),
\bfpage{5170}
(\byear{2019})
\end{barticle}
\endbibitem

\bibitem{binns2018fairness}
\begin{bchapter}
\bauthor{\bsnm{Binns}, \binits{R.}}:
\bctitle{Fairness in machine learning: Lessons from political philosophy}.
In: \bbtitle{Conference on Fairness, Accountability and Transparency},
pp. \bfpage{149}--\blpage{159}
(\byear{2018}).
\bcomment{PMLR}
\end{bchapter}
\endbibitem

\bibitem{raji2019actionable}
\begin{bchapter}
\bauthor{\bsnm{Raji}, \binits{I.D.}},
\bauthor{\bsnm{Buolamwini}, \binits{J.}}:
\bctitle{Actionable auditing: Investigating the impact of publicly naming biased performance results of commercial ai products}.
In: \bbtitle{Proceedings of the 2019 AAAI/ACM Conference on AI, Ethics, and Society},
pp. \bfpage{429}--\blpage{435}
(\byear{2019})
\end{bchapter}
\endbibitem

\bibitem{messeri2024artificial}
\begin{barticle}
\bauthor{\bsnm{Messeri}, \binits{L.}},
\bauthor{\bsnm{Crockett}, \binits{M.}}:
\batitle{Artificial intelligence and illusions of understanding in scientific research}.
\bjtitle{Nature}
\bvolume{627}(\bissue{8002}),
\bfpage{49}--\blpage{58}
(\byear{2024})
\end{barticle}
\endbibitem

\bibitem{holstein2019improving}
\begin{bchapter}
\bauthor{\bsnm{Holstein}, \binits{K.}},
\bauthor{\bsnm{Wortman~Vaughan}, \binits{J.}},
\bauthor{\bsnm{Daum{\'e}~III}, \binits{H.}},
\bauthor{\bsnm{Dudik}, \binits{M.}},
\bauthor{\bsnm{Wallach}, \binits{H.}}:
\bctitle{Improving fairness in machine learning systems: What do industry practitioners need?}
In: \bbtitle{Proceedings of the 2019 CHI Conference on Human Factors in Computing Systems},
pp. \bfpage{1}--\blpage{16}
(\byear{2019})
\end{bchapter}
\endbibitem

\bibitem{schwartz2021proposal}
\begin{botherref}
\oauthor{\bsnm{Schwartz}, \binits{R.}},
\oauthor{\bsnm{Down}, \binits{L.}},
\oauthor{\bsnm{Jonas}, \binits{A.}},
\oauthor{\bsnm{Tabassi}, \binits{E.}}:
A proposal for identifying and managing bias in artificial intelligence.
Draft NIST Special Publication
\textbf{1270}
(2021)
\end{botherref}
\endbibitem

\bibitem{schwartz2022towards}
\begin{bbook}
\bauthor{\bsnm{Schwartz}, \binits{R.}},
\bauthor{\bsnm{Schwartz}, \binits{R.}},
\bauthor{\bsnm{Vassilev}, \binits{A.}},
\bauthor{\bsnm{Greene}, \binits{K.}},
\bauthor{\bsnm{Perine}, \binits{L.}},
\bauthor{\bsnm{Burt}, \binits{A.}},
\bauthor{\bsnm{Hall}, \binits{P.}}:
\bbtitle{Towards a Standard for Identifying and Managing Bias in Artificial Intelligence}
vol. \bseriesno{3}.
\bpublisher{US Department of Commerce, National Institute of Standards and Technology}, \blocation{???}
(\byear{2022})
\end{bbook}
\endbibitem

\bibitem{van2023ai}
\begin{barticle}
\bauthor{\bsnm{Van~Noorden}, \binits{R.}},
\bauthor{\bsnm{Perkel}, \binits{J.M.}}:
\batitle{Ai and science: what 1,600 researchers think}.
\bjtitle{Nature}
\bvolume{621}(\bissue{7980}),
\bfpage{672}--\blpage{675}
(\byear{2023})
\end{barticle}
\endbibitem

\bibitem{gorriz2020artificial}
\begin{barticle}
\bauthor{\bsnm{G{\'o}rriz}, \binits{J.M.}},
\bauthor{\bsnm{Ram{\'\i}rez}, \binits{J.}},
\bauthor{\bsnm{Ortiz}, \binits{A.}},
\bauthor{\bsnm{Martinez-Murcia}, \binits{F.J.}},
\bauthor{\bsnm{Segovia}, \binits{F.}},
\bauthor{\bsnm{Suckling}, \binits{J.}},
\bauthor{\bsnm{Leming}, \binits{M.}},
\bauthor{\bsnm{Zhang}, \binits{Y.-D.}},
\bauthor{\bsnm{{\'A}lvarez-S{\'a}nchez}, \binits{J.R.}},
\bauthor{\bsnm{Bologna}, \binits{G.}}, \betal:
\batitle{Artificial intelligence within the interplay between natural and artificial computation: Advances in data science, trends and applications}.
\bjtitle{Neurocomputing}
\bvolume{410},
\bfpage{237}--\blpage{270}
(\byear{2020})
\end{barticle}
\endbibitem

\bibitem{verganti2020innovation}
\begin{barticle}
\bauthor{\bsnm{Verganti}, \binits{R.}},
\bauthor{\bsnm{Vendraminelli}, \binits{L.}},
\bauthor{\bsnm{Iansiti}, \binits{M.}}:
\batitle{Innovation and design in the age of artificial intelligence}.
\bjtitle{Journal of product innovation management}
\bvolume{37}(\bissue{3}),
\bfpage{212}--\blpage{227}
(\byear{2020})
\end{barticle}
\endbibitem

\bibitem{madanchian2024ai}
\begin{botherref}
\oauthor{\bsnm{Madanchian}, \binits{M.}},
\oauthor{\bsnm{Taherdoost}, \binits{H.}}:
Ai-powered innovations in high-tech research and development: From theory to practice.
Computers, Materials \& Continua
\textbf{81}(2)
(2024)
\end{botherref}
\endbibitem

\bibitem{wallach2015atomnet}
\begin{botherref}
\oauthor{\bsnm{Wallach}, \binits{I.}},
\oauthor{\bsnm{Dzamba}, \binits{M.}},
\oauthor{\bsnm{Heifets}, \binits{A.}}:
Atomnet: a deep convolutional neural network for bioactivity prediction in structure-based drug discovery.
arXiv preprint arXiv:1510.02855
(2015)
\end{botherref}
\endbibitem

\bibitem{staszak2022machine}
\begin{barticle}
\bauthor{\bsnm{Staszak}, \binits{M.}},
\bauthor{\bsnm{Staszak}, \binits{K.}},
\bauthor{\bsnm{Wieszczycka}, \binits{K.}},
\bauthor{\bsnm{Bajek}, \binits{A.}},
\bauthor{\bsnm{Roszkowski}, \binits{K.}},
\bauthor{\bsnm{Tylkowski}, \binits{B.}}:
\batitle{Machine learning in drug design: Use of artificial intelligence to explore the chemical structure--biological activity relationship}.
\bjtitle{Wiley Interdisciplinary Reviews: Computational Molecular Science}
\bvolume{12}(\bissue{2}),
\bfpage{1568}
(\byear{2022})
\end{barticle}
\endbibitem

\bibitem{suh2020evolving}
\begin{barticle}
\bauthor{\bsnm{Suh}, \binits{C.}},
\bauthor{\bsnm{Fare}, \binits{C.}},
\bauthor{\bsnm{Warren}, \binits{J.A.}},
\bauthor{\bsnm{Pyzer-Knapp}, \binits{E.O.}}:
\batitle{Evolving the materials genome: How machine learning is fueling the next generation of materials discovery}.
\bjtitle{Annual Review of Materials Research}
\bvolume{50}(\bissue{1}),
\bfpage{1}--\blpage{25}
(\byear{2020})
\end{barticle}
\endbibitem

\bibitem{kim2024materials}
\begin{botherref}
\oauthor{\bsnm{Kim}, \binits{H.}},
\oauthor{\bsnm{Choi}, \binits{H.}},
\oauthor{\bsnm{Kang}, \binits{D.}},
\oauthor{\bsnm{Lee}, \binits{W.B.}},
\oauthor{\bsnm{Na}, \binits{J.}}:
Materials discovery with extreme properties via reinforcement learning-guided combinatorial chemistry.
Chemical Science
(2024)
\end{botherref}
\endbibitem

\bibitem{light2023avalonbench}
\begin{bchapter}
\bauthor{\bsnm{Light}, \binits{J.}},
\bauthor{\bsnm{Cai}, \binits{M.}},
\bauthor{\bsnm{Shen}, \binits{S.}},
\bauthor{\bsnm{Hu}, \binits{Z.}}:
\bctitle{Avalonbench: Evaluating llms playing the game of avalon}.
In: \bbtitle{NeurIPS 2023 Foundation Models for Decision Making Workshop}
(\byear{2023})
\end{bchapter}
\endbibitem

\bibitem{du2024multi}
\begin{botherref}
\oauthor{\bsnm{Du}, \binits{Z.}},
\oauthor{\bsnm{Qian}, \binits{C.}},
\oauthor{\bsnm{Liu}, \binits{W.}},
\oauthor{\bsnm{Xie}, \binits{Z.}},
\oauthor{\bsnm{Wang}, \binits{Y.}},
\oauthor{\bsnm{Dang}, \binits{Y.}},
\oauthor{\bsnm{Chen}, \binits{W.}},
\oauthor{\bsnm{Yang}, \binits{C.}}:
Multi-agent software development through cross-team collaboration.
arXiv preprint arXiv:2406.08979
(2024)
\end{botherref}
\endbibitem

\bibitem{raffel2020exploring}
\begin{barticle}
\bauthor{\bsnm{Raffel}, \binits{C.}},
\bauthor{\bsnm{Shazeer}, \binits{N.}},
\bauthor{\bsnm{Roberts}, \binits{A.}},
\bauthor{\bsnm{Lee}, \binits{K.}},
\bauthor{\bsnm{Narang}, \binits{S.}},
\bauthor{\bsnm{Matena}, \binits{M.}},
\bauthor{\bsnm{Zhou}, \binits{Y.}},
\bauthor{\bsnm{Li}, \binits{W.}},
\bauthor{\bsnm{Liu}, \binits{P.J.}}:
\batitle{Exploring the limits of transfer learning with a unified text-to-text transformer}.
\bjtitle{Journal of machine learning research}
\bvolume{21}(\bissue{140}),
\bfpage{1}--\blpage{67}
(\byear{2020})
\end{barticle}
\endbibitem

\bibitem{openai}
\begin{botherref}
\oauthor{\bsnm{OpenAI}}:
{GPT-4} technical report.
CoRR
(2023)
\end{botherref}
\endbibitem

\bibitem{llama3}
\begin{botherref}
\oauthor{\bsnm{Dubey}, \binits{A.}},
\oauthor{\bsnm{Jauhri}, \binits{A.}},
\oauthor{\bsnm{Pandey}, \binits{A.}},
\oauthor{\bsnm{Kadian}, \binits{A.}},
\oauthor{\bsnm{Al-Dahle}, \binits{A.}},
\oauthor{\bsnm{Letman}, \binits{A.}},
\oauthor{\bsnm{Mathur}, \binits{A.}},
\oauthor{\bsnm{Schelten}, \binits{A.}},
\oauthor{\bsnm{Yang}, \binits{A.}},
\oauthor{\bsnm{Fan}, \binits{A.}}, et al.:
The llama 3 herd of models.
arXiv preprint arXiv:2407.21783
(2024)
\end{botherref}
\endbibitem

\bibitem{kenton2019bert}
\begin{bchapter}
\bauthor{\bsnm{Kenton}, \binits{J.D.M.-W.C.}},
\bauthor{\bsnm{Toutanova}, \binits{L.K.}}:
\bctitle{Bert: Pre-training of deep bidirectional transformers for language understanding}.
In: \bbtitle{Proceedings of naacL-HLT},
vol. \bseriesno{1},
p. \bfpage{2}
(\byear{2019}).
\bcomment{Minneapolis, Minnesota}
\end{bchapter}
\endbibitem

\bibitem{ji2021dnabert}
\begin{barticle}
\bauthor{\bsnm{Ji}, \binits{Y.}},
\bauthor{\bsnm{Zhou}, \binits{Z.}},
\bauthor{\bsnm{Liu}, \binits{H.}},
\bauthor{\bsnm{Davuluri}, \binits{R.V.}}:
\batitle{Dnabert: pre-trained bidirectional encoder representations from transformers model for dna-language in genome}.
\bjtitle{Bioinformatics}
\bvolume{37}(\bissue{15}),
\bfpage{2112}--\blpage{2120}
(\byear{2021})
\end{barticle}
\endbibitem

\bibitem{zhang2024chemllm}
\begin{botherref}
\oauthor{\bsnm{Zhang}, \binits{D.}},
\oauthor{\bsnm{Liu}, \binits{W.}},
\oauthor{\bsnm{Tan}, \binits{Q.}},
\oauthor{\bsnm{Chen}, \binits{J.}},
\oauthor{\bsnm{Yan}, \binits{H.}},
\oauthor{\bsnm{Yan}, \binits{Y.}},
\oauthor{\bsnm{Li}, \binits{J.}},
\oauthor{\bsnm{Huang}, \binits{W.}},
\oauthor{\bsnm{Yue}, \binits{X.}},
\oauthor{\bsnm{Zhou}, \binits{D.}}, et al.:
Chemllm: A chemical large language model.
arXiv preprint arXiv:2402.06852
(2024)
\end{botherref}
\endbibitem

\bibitem{liu2024benchmarking}
\begin{botherref}
\oauthor{\bsnm{Liu}, \binits{J.}},
\oauthor{\bsnm{Zhou}, \binits{P.}},
\oauthor{\bsnm{Hua}, \binits{Y.}},
\oauthor{\bsnm{Chong}, \binits{D.}},
\oauthor{\bsnm{Tian}, \binits{Z.}},
\oauthor{\bsnm{Liu}, \binits{A.}},
\oauthor{\bsnm{Wang}, \binits{H.}},
\oauthor{\bsnm{You}, \binits{C.}},
\oauthor{\bsnm{Guo}, \binits{Z.}},
\oauthor{\bsnm{Zhu}, \binits{L.}}, et al.:
Benchmarking large language models on cmexam-a comprehensive chinese medical exam dataset.
Advances in Neural Information Processing Systems
\textbf{36}
(2024)
\end{botherref}
\endbibitem

\bibitem{zhang2023exploring}
\begin{bchapter}
\bauthor{\bsnm{Zhang}, \binits{J.}},
\bauthor{\bsnm{Xu}, \binits{X.}},
\bauthor{\bsnm{Zhang}, \binits{N.}},
\bauthor{\bsnm{Liu}, \binits{R.}},
\bauthor{\bsnm{Hooi}, \binits{B.}},
\bauthor{\bsnm{Deng}, \binits{S.}}:
\bctitle{Exploring collaboration mechanisms for {LLM} agents: {A} social psychology view}.
In: \bbtitle{Proceedings of the 62nd Annual Meeting of the Association for Computational Linguistics},
pp. \bfpage{14544}--\blpage{14607}
(\byear{2024})
\end{bchapter}
\endbibitem

\bibitem{nogueira2019passage}
\begin{botherref}
\oauthor{\bsnm{Nogueira}, \binits{R.}},
\oauthor{\bsnm{Cho}, \binits{K.}}:
Passage re-ranking with bert.
arXiv preprint arXiv:1901.04085
(2019)
\end{botherref}
\endbibitem

\bibitem{mars2022word}
\begin{barticle}
\bauthor{\bsnm{Mars}, \binits{M.}}:
\batitle{From word embeddings to pre-trained language models: A state-of-the-art walkthrough}.
\bjtitle{Applied Sciences}
\bvolume{12}(\bissue{17}),
\bfpage{8805}
(\byear{2022})
\end{barticle}
\endbibitem

\bibitem{alsayat2022improving}
\begin{barticle}
\bauthor{\bsnm{Alsayat}, \binits{A.}}:
\batitle{Improving sentiment analysis for social media applications using an ensemble deep learning language model}.
\bjtitle{Arabian Journal for Science and Engineering}
\bvolume{47}(\bissue{2}),
\bfpage{2499}--\blpage{2511}
(\byear{2022})
\end{barticle}
\endbibitem

\bibitem{petukhova2024text}
\begin{botherref}
\oauthor{\bsnm{Petukhova}, \binits{A.}},
\oauthor{\bsnm{Matos-Carvalho}, \binits{J.P.}},
\oauthor{\bsnm{Fachada}, \binits{N.}}:
Text clustering with llm embeddings.
arXiv preprint arXiv:2403.15112
(2024)
\end{botherref}
\endbibitem

\end{thebibliography}

\clearpage

\appendix
\renewcommand{\contentsname}{Appendix}

\section{Related Work}
\label{sec:related_work}

\subsection{AI for Science}
In recent years, AI has become increasingly common in science and is expected to become the center of research practice~\citep{van2023ai}. 
AI has demonstrated great potential to accelerate experimental design, data analysis, optimization problem solving, and discovery of new theories~\citep{gorriz2020artificial,verganti2020innovation,madanchian2024ai}.
Specifically, deep neural networks are used to predict the relationship between molecular structures and biological activity~\citep{wallach2015atomnet,staszak2022machine}, reinforcement learning is used to discover unknown materials with superior properties~\citep{suh2020evolving,kim2024materials}, and agent-based systems are introduced to simulate social science scenarios~\citep{light2023avalonbench,du2024multi}.
In addition, as a subfield of science, AI has undergone some preliminary explorations in the SoS~\citep{alshebli2018preeminence,shi2023surprising,su2024two}, revealing promising results.

\subsection{Large Language Models}

The role of large language models (LLMs) can be articulated from two perspectives: chat (T5~\citep{raffel2020exploring}, GPT-4~\citep{openai}, and LLaMA3.1~\citep{llama3}) and embedding (BERT~\citep{kenton2019bert} and DNABERT~\citep{ji2021dnabert}) generation. 
First, the capability of dialogue generation enables LLMs to understand user input in natural language and generate contextually relevant responses in various conversational contexts such as knowledge testing, game play, and software programming~\citep{zhang2024chemllm,liu2024benchmarking,zhang2023exploring,du2024multi}. 
Additionally, embedding generation allows LLMs to convert input text into fixed-dimensional vector representations, which effectively capture the semantic information of the text and can be used for tasks such as text similarity computation, information retrieval, and sentiment analysis~\citep{nogueira2019passage,mars2022word,alsayat2022improving,petukhova2024text}.
Therefore, the capabilities of LLMs in both text generation and embedding generation make them applications spanning from natural language processing tasks to more complex domains such as SoS, where they can assist in understanding research dynamics, scientific discovery, and scientific collaboration.


\section{Review and Indexing System}\label{sec:review}

In Table~\ref{tab:prompt_review_1} and~\ref{tab:prompt_review_2}, we present the peer review criteria used in our simulation system, which is based on the modified Neural Information Processing Systems review guidelines~\footnote{\url{https://neurips.cc/Conferences/2024/ReviewerGuidelines}} considering that the papers produced by cross-discipline agents are not all in the field of computer science. Although this criteria comes from a computer science conference, the basic evaluation metrics can be applied in multiple areas. 
\begin{table*}[ht]
\centering
\caption{Prompt Tailored for Multidisciplinary Reviewers}
\label{tab:prompt_review_1}
\resizebox{0.99\textwidth}{!}{
\begin{tabularx}{\textwidth}{X}
\toprule
\textbf{Prompt Tailored for Multidisciplinary Reviewers (1/2)} \\
\midrule
You are a researcher from a multidisciplinary background reviewing a paper that has been submitted to a venue that involves multiple scientific disciplines. Be critical and cautious in your decision-making. If the paper has significant weaknesses or you are uncertain about its quality, provide lower scores and recommend rejection. Below are the questions you will be asked on the review form for each paper and some guidelines on what to consider when answering these questions.

Reviewer Guidelines for Multidisciplinary Paper Review:

1. Summary:  
Provide a brief summary of the paper and its contributions. This is not the place to critique the paper. The authors should generally agree with a well-written summary, which reflects an accurate understanding of their work from a multidisciplinary perspective.

2. Strengths and Weaknesses:  
Please provide a thorough assessment of the strengths and weaknesses of the paper, touching on each of the following dimensions:

\quad - Originality: Are the tasks or methods novel within each of the relevant disciplines? Does the work represent an innovative combination of techniques or concepts from different fields? Is it clear how this work distinguishes itself from previous contributions in each discipline involved?

\quad - Quality: Is the submission technically sound in each of the relevant fields? Are claims well-supported by evidence (e.g., theoretical analysis or experimental results)? Are the methods used appropriately for each discipline involved? Is this a complete piece of work, or still a work in progress? Are the authors transparent and honest in evaluating both the strengths and weaknesses of their work?

\quad - Clarity: Is the paper written in a way that is accessible to readers from multiple disciplines? Is it well-organized, with clear explanations of concepts across different fields? If not, please suggest improvements for clarity. Does it provide sufficient detail for an expert in each relevant field to understand the methodology and reproduce results?

\quad - Significance: Are the results important? Are others (researchers or practitioners) likely to use the ideas or build on them? Does the submission address a difficult task in a better way than previous work? Does it advance the state of the art in a demonstrable way? Does it provide unique data, unique conclusions about existing data, or a unique theoretical or experimental approach?

3. Questions:
Please list any questions or suggestions that could help clarify the paper’s limitations or improve its quality. Responses from the authors could change your opinion or address areas of confusion. This feedback can be critical for the rebuttal and discussion phase with the authors.\\

\bottomrule
\end{tabularx}}
\end{table*}

\begin{table*}[ht]
\centering
\caption{Prompt Tailored for Multidisciplinary Reviewers}
\label{tab:prompt_review_2}
\resizebox{0.99\textwidth}{!}{
\begin{tabularx}{\textwidth}{X}
\toprule
\textbf{Prompt Tailored for Multidisciplinary Reviewers (2/2)} \\
\midrule
4. Ethical Concerns: 
Flag any ethical concerns, particularly those that may arise from interdisciplinary collaboration. Ensure any ethical issues related to research design, data usage, or broader implications are addressed.

5. Overall Score: 
Provide a final score based on the paper’s strengths and weaknesses. Use the following scale:

\quad - 10: Award Quality:  
  A technically flawless paper with groundbreaking impact across one or more disciplines, with exceptionally strong evaluation, reproducibility, and resources, and no unaddressed ethical concerns.

\quad - 9: Very Strong Accept: 
  A technically flawless paper with groundbreaking impact in at least one area and strong impact on multiple areas, with flawless evaluation, resources, and reproducibility, and no unaddressed ethical concerns.

\quad - 8: Strong Accept:  
  A technically strong paper with novel ideas, significant impact on at least one discipline or moderate-to-high impact on multiple areas, with excellent evaluation, resources, and reproducibility, and no unaddressed ethical concerns.

\quad - 7: Accept:  
  A technically solid paper with moderate-to-high impact in one or more subfields, good-to-excellent evaluation, reproducibility, and resources, and no unaddressed ethical concerns.

\quad - 6: Weak Accept:  
  A solid paper with moderate impact, no major concerns in terms of evaluation, reproducibility, and ethical considerations.

\quad - 5: Borderline Accept:  
  A technically solid paper where reasons to accept outweigh reasons to reject, e.g., limited evaluation. Use sparingly.

\quad - 4: Borderline Reject:  
  A technically solid paper where reasons to reject outweigh reasons to accept, e.g., limited evaluation. Use sparingly.

\quad - 3: Reject:  
  A paper with technical flaws, weak evaluation, inadequate reproducibility, or incompletely addressed ethical concerns.

\quad - 2: Strong Reject:  
  A paper with major technical flaws, poor evaluation, limited impact, poor reproducibility, or mostly unaddressed ethical considerations.

\quad - 1: Very Strong Reject:  
  A paper with trivial results, poor evaluation, or unaddressed ethical issues.  \\
\bottomrule
\end{tabularx}}
\end{table*}

\end{document}